# Dimension Reduction with Prior Information for Knowledge Discovery

Anh Tuan Bui

**Abstract**— This paper addresses the problem of mapping high-dimensional data to a low-dimensional space, in the presence of other known features. This problem is ubiquitous in science and engineering as there are often controllable/measurable features in most applications. To solve this problem, this paper proposes a broad class of methods, which is referred to as conditional multidimensional scaling (MDS). An algorithm for optimizing the objective function of conditional MDS is also developed. The convergence of this algorithm is proven under mild assumptions. Conditional MDS is illustrated with kinship terms, facial expressions, textile fabrics, car-brand perception, and cylinder machining examples. These examples demonstrate the advantages of conditional MDS over conventional dimension reduction in improving the estimation quality of the reduced-dimension space and simplifying visualization and knowledge discovery tasks. Computer codes for this work are available in the open-source **cml** R package.

## 1 Introduction

Dimension reduction is necessary for scientists to discover hidden features from high-dimensional data. Specifically, dimension reduction methods meaningfully map high-dimensional observations to some low-dimensional space. Ideally, the reduced dimension should be as small as the intrinsic dimension of the data [1]. To achieve this goal, principal component analysis (PCA) and factor analysis are widely used for learning a linear map. For nonlinear cases, kernel PCA [2] and manifold learning [3] are among popular choices. Autoencoder is another nonlinear dimension reduction method that has been successfully applied in many studies (see, e.g., [4], [5]).

Most of dimension reduction methods require input data to be in the vector form. However, data are sometimes in dissimilarity/similarity forms (e.g., similarity rating or intercorrelation). In many applications, practitioners may also want to define suitable pairwise dissimilarity measures for their data, such as dissimilarity indices for community ecology data [6], dissimilarity for image data of random heterogeneous materials [7], dissimilarity between unstructured point cloud data [8], the cosine dissimilarity commonly used for text data [9], and distance measures for statistical distributions [10]. This paper develops a dimension reduction method that directly takes pairwise dissimilarities as the input.

Existing dissimilarity-based dimension reduction methods (see Section 2.1 for a brief review) largely cannot incorporate existing knowledge of known features into the dimension reduction process. For instance, Section 4.5 discusses a cylinder machining example, in which the cylinders are machined with different combinations of lathe-turning parameters. Indeed, there are often such controllable and/or measurable features in most science and engineering applications. This limitation generally leads to two potential problems. First, failing to utilize all available information generally leads to poorer estimation of the reduced-dimension space. Second, the known features may mask other features in visualization (due to the influence of dimension arrangement [11], dimension order [12], or visualization distortion [13]); this can hamper visualization and knowledge discovery tasks.

In light of this, this paper proposes a novel dissimilarity-based dimension reduction framework, called *conditional MDS*. Additionally, an optimization algorithm is developed to solve the learning problem of conditional MDS. This algorithm is shown to converge under mild assumptions. The main contribution of this paper is that the proposed conditional MDS methodology and theory endow MDS analysis with the ability to incorporate available knowledge of the known features. This leads to the following advantages:

- Improving the estimation quality of the reduced-dimension space by utilizing more fully available data.
- Simplifying visualization and knowledge discovery tasks in two ways. First, marginalizing the known features out of the reduced-dimension space helps avoid the aforementioned masking problem. Second, the discovered features in previous analyses can be used as the known features in subsequent analyses. This knowledge discovery process is more straightforward because instead of having to recognize all unknown (including unanticipated) features at once, practitioners can now focus on one or a few features at a time.

The organization of this paper is as follows. Section 2 reviews related works, and Section 3 presents the proposed method. Section 4 shows the advantages of the proposed method over existing dissimilarity-based dimension reduction methods via five examples. Discussions on a number of issues of the proposed method are in Section 5. Finally, Section 6 concludes the paper.

*A.T. Bui is with the Department of Statistical Sciences and Operations Research, Virginia Commonwealth University, Richmond, VA 23284. E-mail: buiat2@vcu.edu.*

# 2 Related Works

## 2.1. Dissimilarity-Based Dimension Reduction

Dissimilarity-based dimension reduction methods directly take pairwise dissimilarities as the input. Specifically, given a $N \times N$ pairwise dissimilarity matrix $\boldsymbol{\Delta} = [\delta_{ij}]_{i,j=1,..,N}$ of $N$ objects, their goal is to find low-dimensional embeddings $\{\mathbf{z}_1, \mathbf{z}_2, \dots, \mathbf{z}_N\}$ of these $N$ objects.

Classical MDS [14], Sammon mapping [15], and curvilinear component analysis [16] are among the earliest dissimilarity-based dimension reduction techniques. These methods can be considered as instances of a broad class of metric MDS techniques [17]. Given $\boldsymbol{\Delta} = [\delta_{ij}]_{i,j=1,..,N}$, metric MDS finds $\{\mathbf{z}_1, \mathbf{z}_2, \dots, \mathbf{z}_N\}$ via minimizing the *stress* function:

$$\min_{\mathbf{z}_1,\mathbf{z}_2,\dots,\mathbf{z}_N} \sum_{i<j} w_{ij}\left\{||\mathbf{z}_i - \mathbf{z}_j|| - \tilde{\delta}_{ij}\right\}^2, \quad (1)$$

where $w_{ij}$ is a nonnegative weight, $||\mathbf{z}_i - \mathbf{z}_j||$ is the Euclidean distance between $\mathbf{z}_i$ and $\mathbf{z}_j$, and $\tilde{\delta}_{ij}$ is some non-negative monotonic transformation of $\delta_{ij}$ ($i, j = 1, 2, \dots, N$). The weights are only zero when the corresponding dissimilarities are missing.

Different variants of metric MDS can be derived by changing the weighting and transformation schemes. For instance, classical MDS sets $w_{ij} = 1\ \forall i, j$; Sammon mapping sets $w_{ij} = 1/\left(\tilde{\delta}_{ij} \sum_{i<j} \tilde{\delta}_{ij}\right)\ \forall i, j$ ; and curvilinear component analysis sets the $w_{ij}$'s as some decreasing function of $||\mathbf{z}_i - \mathbf{z}_j||$. All these three methods do not require transformation of the $\delta_{ij}$'s (i.e., $\tilde{\delta}_{ij} = \delta_{ij}\ \forall i, j$). Classical MDS assumes that $\delta_{ij} = ||\mathbf{z}_i - \mathbf{z}_j||\ \forall i, j$, in which case its solution can be found via an eigendecomposition of the "doubly centered" matrix of $\boldsymbol{\Delta}$. This assumption is often violated in practice. Therefore, metric MDS generally relaxes this assumption and finds the solution via iterative optimization. SMACOF [18] is the most popular algorithm for solving the optimization problem of metric MDS via majorization [17].

Metric MDS can be viewed as a class of manifold learning algorithms, which aim to find an implicit manifold representation of the data via the embeddings $\mathbf{z}_i$'s. Sammon mapping and curvilinear component analysis can learn nonlinear manifolds they emphasize the local behavior through the weights $w_{ij}$'s. Classical MDS can only handle linear manifolds because it does not use any weighting or transformation scheme.

More recently, some popular dissimilarity-based dimension reduction approaches include isometric mapping (ISOMAP) [19], t-distributed stochastic neighbor embedding (t-SNE) [20], and uniform manifold approximation and projection (UMAP) [21]. ISOMAP extends classical MDS to handle nonlinear manifolds by transforming the distances $\delta_{ij}$'s to "geodesic" distances. The geodesic distance is defined as the shortest distance between points along the manifold. It can be estimated by forming a neighborhood graph and minimizing the sum of the Euclidean distances on this graph. ISOMAP obtains the embeddings $\mathbf{z}_i$'s via applying classical MDS to the geodesic distances.

Unlike MDS-based approaches, which focus on preserving dissimilarities, t-SNE aims to conserve probabilities. In particular, t-SNE requires that the conditional probability of picking a point as a neighbor of a given point in the embedding space is similar to that in the data space. To achieve this, t-SNE minimizes the Kullback-Leibler divergence of the probability distribution in the embedding space from that in the data space.

UMAP is another approach that diverges from dissimilarity preservation. Particularly, UMAP tries to maintain topological representations of the embeddings and the data. To construct these representations, UMAP uses fuzzy simplicial set representations of local manifold approximations. It finds the solution via minimizing the cross-entropy of the two fuzzy sets.

Numerous variants of these methods have been proposed in the literature, such as local MDS [22], spectral MDS [23], robust kernel ISOMAP [24], upgraded landmark ISOMAP [25], dynamic t-SNE [26], opt-SNE [27], progressive UMAP [28], and density-preserving UMAP [29]. However, these methods are unable to incorporate available knowledge of the manifolds. In many situations, we already know a subset of the manifold features. Recall the cylinder machining example in Section 4.5, if a conventional dimension reduction method is applied in this example, the lathe-turning parameters used to machine the cylinders will be rediscovered [8]. This is unnecessary and may hinder the discovery of other unknown features. Moreover, failing to utilize the available information of these parameters could adversely affect the learning accuracy of the reduced-dimension space. To address these issues, this paper develops the conditional MDS framework to marginalize the effects of the known features to learn solely the other unknown features.

## 2.2. Supervised Dimension Reduction

A common usage of dimension reduction methods is for mitigating the curse of dimensionality, to build a better predictive model for some outcome based on the reduced-dimension data. In this setting, dimension reduction can be viewed as a preprocessing step. However, by disregarding the outcome, one could expect that this solution is suboptimal.

To address this problem, dimension reduction methods have been extended to consider the outcome. Many such extensions are based on the manifold learning techniques presented in Section 2.1. For example, supervised ISOMAP (S-ISOMAP) [30] transforms the given dissimilarities such that the transformed dissimilarities of intra-class observations are less than that of inter-class observations. Then, ISOMAP is applied to the transformed dissimilarities. Marginal ISOMAP (M-ISOMAP) [31] uses class information to construct constrained neighborhood graphs, in which edges exist if and only if observations are in the same class. This helps avoid several hyper-parameters required by the dissimilarity transformation of S-ISOMAP. Multi-Manifold Discriminant ISOMAP (MMD-ISOMAP) [32] extends M-ISOMAP to the multi-manifold setting (i.e., observations in different classes lie on different manifolds). To this end, MMD-ISOMAP maximizes the inter-class distances in the reduced-dimension space. Semi-supervised local multi-manifold ISOMAP [33] further extends MMD-ISOMAP to utilize both labeled and unlabeled data, via maximizing inter-manifold distances and minimizing intra-manifold distances. Note that these extensions can be readily adopted for MDS techniques. Also, there have been extensions to the supervised setting for t-SNE [34], [35], and UMAP [36]. In addition to supervised dissimilarity-based manifold learning, other supervised dimension reduction methods based on PCA and non-negative matrix factorization can be found in the review paper of Chao et al. [37].

However, supervised dimension reduction solves a fundamentally different problem from the one in this paper. Essentially, supervised dimension reduction assumes that there exists a function that maps the features in the reduced-dimension space to the outcome. If we treat the known features as the outcome, we are assuming that such functional relationships exist between the known and unknown features. However, in the context of this paper, the known features do not necessarily depend on the unknown features. For example, in the kinship terms example in Section 4.1, Gender (say, a known feature) is not a function of Generation (say, an unknown feature).

Likewise, the problem and methods proposed in this paper are not the same as those in the literature of "sufficient dimension reduction". Sufficient dimension reduction approaches (see, e.g., Cook [38] and the references therein) seek for a sufficient statistic **z**, as a function of the given data in a reduced-dimension space, such that the conditional distribution of the outcome given the data is the same as the conditional distribution of the outcome given **z**. Again, this entails a functional relationship between the outcome and **z**.

"Interactive dimension reduction" methods [39], [40] are somewhat related to the proposed conditional MDS framework in the sense that these approaches also incorporate domain knowledge into the dimension reduction process. Nevertheless, a key difference between these methods and conditional MDS is that these methods require practitioners to interact actively with the dimension reduction process. For example, the practitioners adjust features weights, change model parameters, or manipulate outputs, so as to produce comprehensible dimension reduction results. In contrast, there is no direct interaction between conditional MDS and practitioners during the dimension reduction process.

## 3 Conditional Multidimensional Scaling

This section presents the proposed conditional MDS framework and its optimization algorithm. Let $\{\mathbf{u}_i = [u_{i1}, u_{i2}, \ldots, u_{ip}]^T \in \mathbb{R}^p: i = 1, 2, \ldots, N\}$ and $\{\mathbf{v}_i = [v_{i1}, v_{i2}, \ldots, v_{iq}]^T \in \mathbb{R}^q: i = 1, 2, \ldots, N\}$ denote the values of $p$ unknown features and $q$ known features, respectively, of $N$ objects. Note that **u** and **v** constitute all the underlying features that govern the variation in the observed data. Because the intrinsic dimension of the observed data is at most $N-1$, it is required that $N > p + q$. To make the unknown features **u** and the known features **v** compatible in a single manifold coordinate system, an affine transformation $\mathbf{B}^T\mathbf{v}$, where $\mathbf{B} \in \mathbb{R}^{q\times q}$, is needed. Moreover, denote $\mathbf{U}^T = [\mathbf{u}_1, \mathbf{u}_2, \ldots, \mathbf{u}_N] \in \mathbb{R}^{p\times N}$ and $\mathbf{V}^T = [\mathbf{v}_1, \mathbf{v}_2, \ldots, \mathbf{v}_N] \in \mathbb{R}^{q\times N}$. Thus, $[\mathbf{U}, \mathbf{VB}] \in \mathbb{R}^{N\times(p+q)}$ contain all the manifold feature values of the $N$ objects.

The objective function of conditional MDS is referred to as the "conditional stress":

$$\sigma(\mathbf{U},\mathbf{B}) = \sum_{i<j} w_{ij}\left(\tilde{\delta}_{ij} - d_{ij}(\mathbf{U},\mathbf{B})\right)^2, \tag{2}$$

where the $w_{ij}$'s and $\tilde{\delta}_{ij}$'s are as defined in (1), and

$$d_{ij}(\mathbf{U},\mathbf{B}) = \sqrt{\left\|\mathbf{u}_i - \mathbf{u}_j\right\|^2 + \left\|\mathbf{B}^T(\mathbf{v}_i - \mathbf{v}_j)\right\|^2} \tag{3}$$

is the Euclidean distance between the $i$th and $j$th objects in the aforementioned manifold coordinate system for all $i, j$. Conditional MDS minimizes the conditional stress over **U** and **B** to reduce the discrepancy between each pair of the $\tilde{\delta}_{ij}$'s and the $d_{ij}(\mathbf{U},\mathbf{B})$'s. In a sense, this preserves the original information (in the form of dissimilarities) in the reduced-dimension space (in the form of Euclidean distance). Like metric MDS, various variants of the conditional MDS framework can be derived by using different weighting and dissimilarity transformation schemes for the conditional stress function.

In optimization language, the learning task of conditional MDS is solving

$$\mathbf{U}^*, \mathbf{B}^* = \underset{\mathbf{U}\in\mathbb{R}^{N\times p},\mathbf{B}\in\mathbb{R}^{q\times q}}{\operatorname{argmin}} \sigma(\mathbf{U},\mathbf{B}). \tag{4}$$

The SMACOF algorithm [18], which solves the learning problem of metric MDS, does not consider the existence of the known features **v**. It attempts to learn all the known and unknown features altogether. It is inapplicable to solve the learning problem (4) of conditional MDS, which utilizes the information of the known features. Hence, this paper develops a "conditional SMACOF" algorithm to solve (4), based on the majorization technique in SMACOF [18]. In brief, this majorization technique constructs an easy-to-minimize majorizing function for an objective function. In conditional SMACOF, the majorizing function for the conditional stress is quadratic in **U** and **B**. By minimizing this majorizing function, the conditional stress function is also minimized. The update formulas (7) and (8) for **U** and **B**, respectively, in Theorem 1 indeed correspond to a stationary point of this majorizing function. Under Assumption 1, these updates yield a monotonically decreasing sequence of conditional stress values. The proof of Theorem 1 is based on the majorization technique and Lemma 1.

***Theorem 1***: *Define* $\boldsymbol{H} = \left[h_{ij}\right]_{i,j=1}^N \in \mathbb{R}^{N\times N}$ *with*

$$h_{ij} = \begin{cases} -w_{ij} & if\ i \neq j \\ \sum_{k=1,k\neq i}^N w_{ik} & if\ i = j \end{cases} \tag{5}$$

*and let* $\boldsymbol{H}^+$ *be the Moore-Penrose inverses of* $\boldsymbol{H}$*. Furthermore, denote* $\boldsymbol{C}(\boldsymbol{U},\boldsymbol{B}) = \left[c_{ij}\right]_{i,j=1}^N \in \mathbb{R}^{N\times N}$ *with*

$$c_{ij} = \begin{cases} -\frac{w_{ij}\tilde{\delta}_{ij}}{d_{ij}(\boldsymbol{U},\boldsymbol{B})} & if\ d_{ij}(\boldsymbol{U},\boldsymbol{B}) \neq 0 \\ 0 & if\ d_{ij}(\boldsymbol{U},\boldsymbol{B}) = 0 \end{cases} \quad for\ i\neq j, \tag{6}$$

$$and\ c_{ii} = \sum_{j=1,j\neq i}^N c_{ij}.$$

*Additionally, let* $\boldsymbol{U}^{[l]}$ *and* $\boldsymbol{B}^{[l]}$ *denote the values of* $\boldsymbol{U}$ *and* $\boldsymbol{B}$*, respectively, at the* $l^{th}$ *iteration. If* $\boldsymbol{U}^{[l]}$ *and* $\boldsymbol{B}^{[l]}$ *are updated by*

$$\boldsymbol{U}^{[l]} = \boldsymbol{H}^+\boldsymbol{C}(\boldsymbol{U}^{[l-1]},\boldsymbol{B}^{[l-1]})\boldsymbol{U}^{[l-1]} \tag{7}$$

$$\boldsymbol{B}^{[l]} = (\boldsymbol{V}^T\boldsymbol{H}\boldsymbol{V})^{-1}\boldsymbol{V}^T\boldsymbol{C}(\boldsymbol{U}^{[l-1]},\boldsymbol{B}^{[l-1]})\boldsymbol{V}\boldsymbol{B}^{[l-1]} \tag{8}$$

*then under Assumption 1,* $\sigma(\boldsymbol{U}^{[l]},\boldsymbol{B}^{[l]}) \leq \sigma(\boldsymbol{U}^{[l-1]},\boldsymbol{B}^{[l-1]})$*, and the equality occurs when* $\boldsymbol{U}^{[l]} = \boldsymbol{U}^{[l-1]}$ *and* $\boldsymbol{B}^{[l]} = \boldsymbol{B}^{[l-1]}$*.*

***Assumption 1***:

a) *The dissimilarities are symmetric, and thereby, so are the weights $w_{ij}$ (i, j = 1, 2, …, N).*

b) *The weight matrix $[w]_{i,j=1}^{N}$ is irreducible, i.e., it does not contain groups within the N objects for which the intergroup weights are always 0.*

c) ***V** contains q linearly independent difference vectors $\boldsymbol{v}_j - \boldsymbol{v}_k$ $(1 \le j < k \le N)$.*

***Lemma 1**: $\boldsymbol{V}^T\boldsymbol{HV}$ is positive definite under Assumption 1.*

In practice, it is recommended that we stop the iterative updates (7) and (8) when the reduction in the conditional stress is less than some threshold. Note that the conditional stress depends on the scale of the transformed dissimilarities $\tilde{\delta}_{ij}$'s. To provide a scale-invariant measure so that we can use a common threshold for different cases, we define a "normalized conditional stress" by

$$\sigma_n(\mathbf{U},\mathbf{B}) = \frac{\sigma(\mathbf{U},\mathbf{B})}{\sum_{i<j} w_{ij}\tilde{\delta}_{ij}^2} = \frac{\sum_{i<j} w_{ij}\left(\tilde{\delta}_{ij} - d_{ij}(\mathbf{U},\mathbf{B})\right)^2}{\sum_{i<j} w_{ij}\tilde{\delta}_{ij}^2}, \tag{9}$$

Note that the normalized conditional stress represents the proportion of the weighted sum-of-squares of the $\tilde{\delta}_{ij}$'s that is unaccounted for in the reduced-dimension space.

Algorithm 1 summarizes the main steps of the conditional MDS algorithm, based on conditional SMACOF. The required inputs include the given $N \times N$ dissimilarity matrix $\tilde{\Delta} = [\tilde{\delta}_{i,j}]_{i,j=1,\ldots,N}$, the known feature values $\mathbf{V}$, the weights $w_{ij}$'s, the threshold $\gamma$ for min reduction in the normalized conditional stress between consecutive iterations, and the max number of iterations $l_{\max}$.

Step 1 of Algorithm 1 initializes and pre-computes necessary quantities for the iterations in Step 2. **B** can be initialized by an identity matrix, and **U** can be initialized randomly. In Step 2, **U** and **B** are iteratively updated based on (7) and (8), respectively, until the number of iterations reaches $l_{\max}$ or the reduction of the normalized conditional stress $\sigma(\mathbf{U}, \mathbf{B})$ from the previous iteration is not greater than $\gamma$. Finally, Step 3 returns the optimal $\mathbf{U}^*$, $\mathbf{B}^*$, and normalized conditional stress. Note that practitioners may want to run the algorithms a few times to find the initial configuration that yields the smallest normalized conditional stress. The convergence of Algorithm 1 is guaranteed by Corollary 1. All the proofs can be found in the supplementary materials of this paper.

***Corollary 1***: *Under Assumption 1, **U** and **B** will converge to a critical point of the conditional stress function if they are iteratively updated by (7) and (8) as in Algorithm 1 (with $l_{max} = \infty$ and $\gamma = 0$), respectively.*

**Remark 1**: Assumption 1(b) implies that rank(**H**) = **N – 1**. In this case, the Moore-Penrose inverse of **H** can be computed by: $\mathbf{H}^+ = (\mathbf{H} + \mathbf{1}_{N\times N})^{-1} - N^{-2}\mathbf{1}_{N\times N}$, where $\mathbf{1}_{N\times N}$ is an $N \times N$ matrix with all elements equal to 1 [17]. More importantly, it is more computationally efficient to calculate $\mathbf{C}(\mathbf{U}^{[l-1]}, \mathbf{B}^{[l-1]})\mathbf{U}^{[l-1]}$, instead of $\mathbf{H}^+\mathbf{C}(\mathbf{U}^{[l-1]}, \mathbf{B}^{[l-1]})$ in (7) first.

**Algorithm 1: Conditional MDS**

**Inputs:** $\tilde{\Delta}$, **V**, $w_{ij}$ $(\forall i, j)$, $\gamma$, and $l_{\max}$

**Step 1**:
- a) Initialize $\mathbf{U}^{[0]}$ and $\mathbf{B}^{[0]}$
- b) Pre-compute $\mathbf{H}^+$ and $(\mathbf{V}^T\mathbf{HV})^{-1}\mathbf{V}^T$
- c) Compute the initial normalized conditional stress $\sigma_n^{[0]} = \sigma_n^{[0]}(\mathbf{U}^{[0]}, \mathbf{B}^{[0]})$ using (9).
- d) Set $\sigma_n^{[-1]} = \infty$ and $l = 0$

**Step 2:** While $(l < l_{max})$ and $\left(\sigma_n^{[l-1]} - \sigma_n^{[l]} > \gamma\right)$ do:
- a) $l = l + 1$
- b) Update $\mathbf{U}^{[l]}$ and $\mathbf{B}^{[l]}$ by (7) and (8), respectively.
- c) Compute the current normalized conditional stress $\sigma_n^{[l]} = \sigma_n(\mathbf{U}^{[l]}, \mathbf{B}^{[l]})$ using (9).

**Step 3:** Set $\mathbf{U}^* = \mathbf{U}^{[l]}$, $\mathbf{B}^* = \mathbf{B}^{[l]}$, $\sigma_n(\mathbf{U}^*, \mathbf{B}^*) = \sigma_n(\mathbf{U}^{[l]}, \mathbf{B}^{[l]})$

**Outputs:** $\mathbf{U}^*$, $\mathbf{B}^*$, and $\sigma_n(\mathbf{U}^*, \mathbf{B}^*)$

**Remark 2**: If $w_{ij} = 1$ $(\forall i, j = 1, 2, \ldots, N)$, then from (5) we have $\mathbf{H} = N\mathbf{I} - \mathbf{1}_{N\times N}$, where **I** is an $N \times N$ identity matrix. It can be verified that $\mathbf{H}^+ = N^{-2}\mathbf{H} = \mathbf{I}_N/N - \mathbf{1}_{N\times N}/N^2$. This formula avoids having to inverse $(\mathbf{H} + \mathbf{1}_{N\times N})$ and is more computationally efficient for large $N$. Furthermore, the summand $\mathbf{1}_{N\times N}/N^2$ is negligible for large $N$, and we can also approximate $\mathbf{H}^+$ by $\mathbf{H}^+ \approx N^{-1}\mathbf{I}$. This approximation helps avoid not only matrix inversion, but also matrix multiplication (by replacing $\mathbf{H}^+$ in (7) with $N^{-1}$).

**Remark 3**: We may impose a diagonal structure on **B** if we know that the known features contribute to the pairwise distances separately. Denote $b_m$ as the $m$th diagonal element of the diagonal matrix **B** and $b_m^{[l]}$ as the value at the $l$th iteration. Based on (8), $b_m^{[l]}$ $(m = 1, 2, \ldots, q)$ can be updated by

$$b_m^{[l]} = \frac{[\mathbf{V}^T\mathbf{C}(\mathbf{U}^{[l-1]}, \mathbf{B}^{[l-1]})V]_{mm}}{[\mathbf{V}^T\mathbf{HV}]_{mm}} b_m^{[l-1]}, \tag{10}$$

where $[\cdot]_{mm}$ denote the $m$th diagonal element of the matrix inside the square brackets.

**Remark 4**: Supervised/sufficient dimension reduction methods are inapplicable for the problem in this paper even if there exists a functional relationship between the known and unknown features, say **v** = **g**(**u**). Treating **v** as the outcome to use supervised/sufficient dimension reduction will result in learning (implicitly) the function **g**(•), which is different from the underlying function that maps **u** and **v** to the manifold embedded in the data space, say **f**(**u**, **v**) = **f**(**u**, **g**(**u**)).

**Remark 5**: Note in (3) that **B** accounts for the joint effect of the known features **v** on the pairwise distances, in compliance with the unknown features **u**. Therefore, by learning **B**, we can separate the joint effect of the known features on the pairwise distances, from that of the unknown features. This is how conditional MDS can marginalize the known features in the dimension reduction result.

## 4 EXAMPLES

This section demonstrates the advantages of conditional MDS via a number of examples. Specifically, Section 4.1 illustrates how conditional MDS can simplify visualization and knowledge discovery tasks via a kinship terms example. To demonstrate the benefit of incorporating available knowledge of the known features for improving estimation accuracy of the reduced-dimension space, Sections 4.2 – 4.5 compare conditional MDS with four dissimilarity-based dimension reduction methods (metric MDS, ISOMAP, t-SNE, and UMAP) in four examples (facial expressions, textile fabrics, simulated car-brand perception, and cylinder machining). As mentioned in the introduction, we assume that the input data are pairwise dissimilarities. As such, dimension reduction methods that cannot take pairwise dissimilarities as the input (e.g., PCA and its variants) are not considered here.

## 4.1. Kinship Terms

This section demonstrates the advantages of conditional MDS in simplifying visualization and knowledge discovery tasks. As mentioned in Section 1, known features can mask unknown features in visualization. To illustrate this problem, Fig. 1(a) plots the two-dimensional (2D) coordinates found by applying metric multidimensional scaling (MDS) to the pairwise dissimilarities of 14 kinship terms in the study of Rosenberg and Kim [41]. Up to some rotation, a feature can be inferred to be Gender (see Borg et al. [42] for a full treatment of MDS result interpretation). Nevertheless, the interpretation for the other feature in Fig. 1(a) is not obvious because Gender may mask this feature.

Conditional MDS addresses this problem by marginalizing the known features out of the dimension reduction results, to expose unknown features. To see this, Fig. 1(b) shows a similar plot to Fig. 1(a), but for the dimension reduction result of conditional MDS, using Gender as the known feature for the kinship terms. Interestingly, we can see in Fig. 1(b) seven pairs of highly similar kinship terms: Sister/Brother (i.e., Sibling), Mother/Father (i.e., Parent), Daughter/Son (i.e., Child), Grandmother/Grandfather (i.e., Grandparent), Granddaughter/Grandson (i.e., Grandchild), Niece/Nephew (i.e., Nibling), and Aunt/Uncle (i.e., Pibling). With Gender marginalized out of the reduced-dimension space, Fig. 1(b) reveals a change in the kinship degree level along the dashed arrow. In this example, conditional MDS helps identify a new "Kinship Degree" feature.

Moreover, conditional MDS allows practitioners to add the discovered features in previous analyses to the known feature set in subsequent analyses, repeatedly. For example, Fig. 1(c) shows a similar plot to Fig. 1(b), but for the dimension reduction result of conditional MDS using Gender and Kinship Degree as the known features. With both Gender and Kinship Degree marginalized out of the reduced-dimension space, Fig. 1(c) exposes a generation transition (Grandparent → Parent/Pibling → Sibling → Child/Nibling → Grandchild) along the dashed curved arrow. Thus, conditional MDS helps identify additionally a new "Generation" feature.

**Remark 6**: Conditional MDS is not limited for use only with 2D scatterplots as in Fig. 1. One can also use conditional MDS in conjunction with other higher-dimensional visualization techniques (e.g., RadViz plots [43] or star coordinates [44]) to discover several features at once. For illustration, Fig. S1 in the supplementary materials of this paper shows a RadViz plot of the three metric MDS coordinates learned for the 14 kinship terms. This plot reveals the Gender and Kinship Degree features, but not Generation. Nevertheless, a further round of conditional MDS analysis using Gender and Kinship Degree as the known features can expose Generation as in Fig. 1(c). For ease of comprehension for general readers and practitioners, this paper uses 2D scatterplots.

## 4.2. Facial Expressions

In this section, conditional MDS is tested with the facial expressions data in the study of Abelson and Sermat [45]. In this work, the authors obtained pairwise dissimilarities for 13 facial expressions (by asking 30 women to give nine-point dissimilarity ratings of each pair of 13 photographs). They showed that the estimates by Engen et al. [46] of the three Schlosberg scales (Pleasant-Unpleasant (PU), Attention-Rejection (AR), and Tension-Sleep (TS)) for these 13 facial expressions accounted for 75% of the variation in the dissimilarities. Note that the absolute correlations between PU-AR, PU-TS, and AR-TS computed from the data by Engen et al. [46] are 0.18, 0.15, and 0.75, respectively. To validate the performance of conditional MDS, one or two of these Schlosberg scales (PU, AR, and TS) is used as the known feature(s). We treat the other scale(s) as unknown for evaluation purposes. In total, this example considers six different scenarios of the known feature sets.

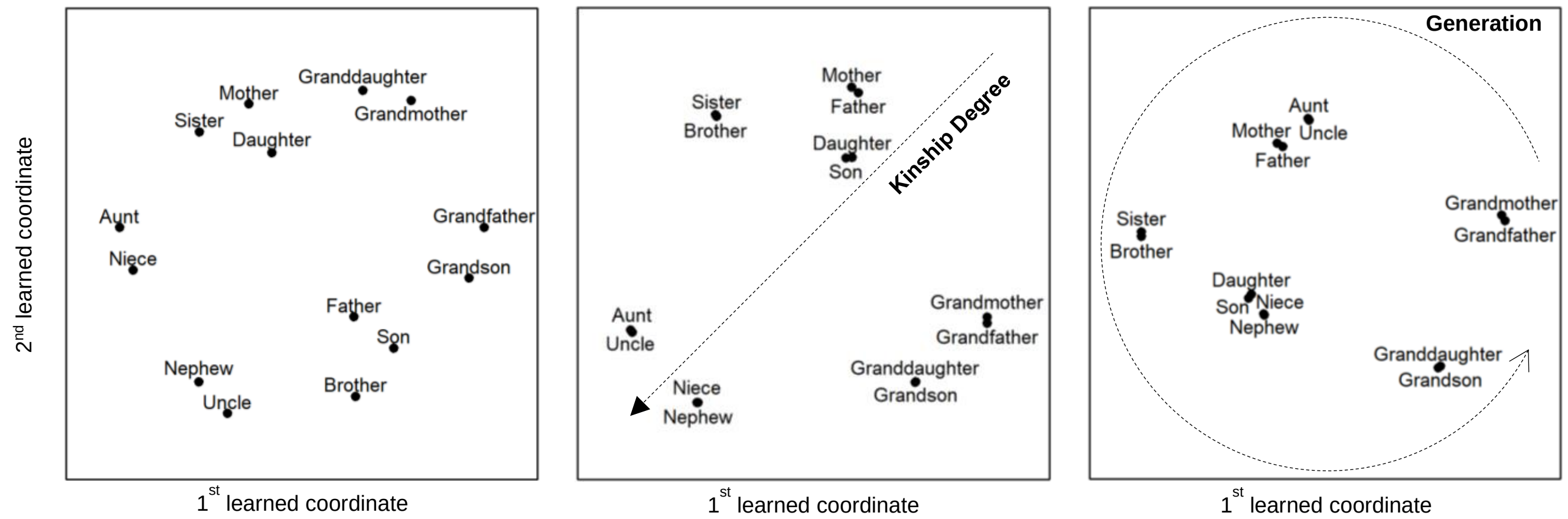

Fig. 1. Scatterplot of the 2D coordinates of 14 kinship terms found by: (a) metric MDS, (b) conditional MDS using Gender as the known feature, and (c) conditional MDS, using Gender and Kinship Degree as the known features. In Fig. 1(a), up to some rotation, a feature is apparently Gender, but the other is unclear as it is possibly a combination of several features. In Fig. 1(b), conditional MDS marginalizes Gender out of the dimension reduction result, revealing a change in the kinship degree level along the dashed arrow. In Fig. 1(c), conditional MDS marginalizes both Gender and Kinship Degree out of the dimension reduction result, exposing a generation transition along the dashed curved arrow.

Conditional MDS is compared with metric MDS based on the average canonical correlation (ACC), as follows. For conditional MDS, we first combine the known feature(s) with the learned feature(s) to have three features in total for each scenario. Then, we compute their ACC with the unknown feature(s). The rationale for this calculation is that the reduced-dimension space of conditional MDS can be viewed to consist of the $p$ learned features and the $q$ known features. For metric MDS, we compute the ACC between the three metric MDS dimensions and all three Schlosberg scales, because metric MDS treats all these scales as unknown. Table 1 reports the ACCs of metric MDS (Column 1) and of conditional MDS for six different scenarios of the known feature set (Columns 2−7). It can be seen from Table 1 that conditional MDS significantly improves the estimation accuracy of the reduced-dimension space in all scenarios.

TABLE 1. AVERAGE CANONICAL CORRELATIONS BETWEEN THE UNKNOWN FEATURES AND THE REDUCED-DIMENSION SPACE LEARNED FOR THE FACIAL EXPRESSIONS BY METRIC MDS AND CONDITIONAL MDS (FOR 6 SCENARIOS OF KNOWN FEATURES).

| Metric | Conditional MDS | | | | | |
|---|---|---|---|---|---|---|
| MDS | {PU} | {AR} | {TS} | {PU, AR} | {PU, TS} | {AR, TS} |
| 0.71 | 0.85 | 0.94 | 0.91 | 0.89 | 0.91 | 0.94 |

To gain more insight of conditional MDS, Table 2 reports the absolute correlations between the first conditional MDS dimension and PU, AR, or TS, for six different scenarios of the known feature set. The results in all scenarios consistently show that the first conditional MDS dimension has weak correlations with the known features and most of the time strong correlations with the unknown features. This suggests that conditional MDS effectively marginalizes the known feature(s) to learn correctly the unknown feature(s). Note that when using PU as the single known feature, in addition to a strong correlation (0.91) between the first conditional MDS dimension and TS, this dimension also exhibits somewhat high correlation with AR (0.72). This suggests that this dimension is related to both TS and AR, which is due to their inherent correlation. The moderate correlations of the first dimension of conditional MDS with the unknown features when using {PU, AR} or {PU, TS} as the known feature set are also likely due to the inherent strong correlation between AR and TS.

### 4.3. Textile Fabrics

This section tests conditional MDS on textile fabric images from the **textile2** dataset [47]. This dataset contains images taken on non-overlapped areas of two fabric samples. These two fabric samples have slightly different textures. We randomly select 15 images from each sample. For evaluation purposes, we randomly apply the following three changes to these 30 images. The first is a 90-degree rotation of the images. The other two are digital contractions of the images in the horizontal and vertical directions by up to 50%. We use the Bernoulli distribution to determine if an image is rotated. And we use the uniform distribution for the amounts of contractions. The resulting images are then cropped to be of size 250×250 pixels. Thus, four underlying features govern the differences among the images: texture, rotation, and vertical/horizontal contractions.

TABLE 2. ABSOLUTE CORRELATIONS BETWEEN THE FIRST CONDITIONAL MDS DIMENSION FOUND FOR THE FACIAL EXPRESSIONS AND PU, AR, OR TS, FOR 6 SCENARIOS OF KNOWN FEATURES. THE BOLD NUMBERS INDICATE THE CORRELATIONS WITH THE UNKNOWN FEATURE(S).

| Known feature set | PU | AR | TS |
|---|---|---|---|
| {PU} | 0.18 | **0.72** | **0.91** |
| {AR} | **0.93** | 0.02 | **0.31** |
| {TS} | **0.92** | **0.14** | 0.15 |
| {PU, AR} | 0.07 | 0.04 | **0.38** |
| {PU, TS} | 0.03 | **0.52** | 0.13 |
| {AR, TS} | **0.93** | 0.07 | 0.16 |

We compute the pairwise dissimilarities of these 30 images by two different measures. The first is the Kullback-Leibler (KL) dissimilarity measure for surface images with random textures of Bui and Apley [7]. The second is the Euclidean distance (ED) when we treat the collection of pixel intensities of each image as a vector of 250 × 250 = 62,500 dimensions. We apply conditional MDS to these dissimilarities, using texture and rotation as the known features. We set the number of unknown features $p = 2$, which is found via the scree-plot-based approach discussed in Section 5.2 (see Fig. S2 in the supplementary materials of this paper). We compare conditional MDS with metric MDS in the same manner as in Section 4.2, but additionally along with ISOMAP, t-SNE, and UMAP.

For conditional MDS, we compute the ACC of the four underlying features with the combined set of the two known and two learned features by conditional MDS. For the metric MDS, ISOMAP, t-SNE, and UMAP, we compute the ACCs between the four underlying features and the four learned features by these methods. Table 3 shows the ACCs of all methods for both KL dissimilarity and ED. It can be seen from Table 3 that when using the KL dissimilarity measure, conditional MDS almost produces perfect canonical correlations between its learned four-dimensional space and the four underlying features. This performance of all other dimension reduction methods are much poorer because they do not utilize the available information of the known features. Furthermore, the performances of all methods are poor when using ED to calculate the dissimilarities. This is due to the random nature of textile fabric textures, which ED cannot capture completely. The stark contrast between ED and KL here highlights the need for dissimilarity-based dimension reduction, so that suitable dissimilarity measures can be used.

TABLE 3. AVERAGE CANONICAL CORRELATIONS BETWEEN THE UNKNOWN FEATURES AND THE FOUR-DIMENSIONAL SPACE LEARNED FOR THE FABRIC IMAGES BY CONDITIONAL MDS AND FOUR DIMENSION REDUCTION METHODS.

| | Metric MDS | ISO-MAP | t-SNE | UMAP | Conditional MDS |
|---|---|---|---|---|---|
| KL | 0.66 | 0.65 | 0.65 | 0.58 | 0.99 |
| ED | 0.24 | 0.26 | 0.27 | 0.30 | 0.61 |

To show more insights of conditional MDS, Table 4 reports the correlation of the two learned dimensions by conditional MDS from the KL dissimilarities with the four underlying features. Again, the learned dimensions are strongly correlated with at least one unknown feature (horizontal/vertical contraction) and weakly correlated with the known features (texture and rotation).

TABLE 4. CORRELATIONS BETWEEN THE TWO LEARNED DIMENSIONS BY CONDITIONAL MDS WITH THE FOUR UNDERLYING FEATURES IN THE TEXTILE FABRICS EXAMPLE.

| | Horizontal contraction | Vertical contraction | Rotation | Texture |
|---|---|---|---|---|
| 1st dim. | 0.43 | 0.87 | 0.05 | 0.14 |
| 2nd dim. | 0.86 | 0.46 | 0.03 | 0.08 |

### 4.4. Car-Brand Perception

This section evaluates conditional MDS with a simulated car-brand perception example. Suppose that the seven features in Table 5 contribute to the dissimilarities between car brands. The Uniform(0, 1) distribution is used to generate the values of these features for $N$ = 30 car brands. Then, the pairwise dissimilarities between car brands are calculated as the weighted Euclidean distances of the feature vectors (using the weights in Table 5 divided by their sum), plus normal noises with zero means and standard deviations equal to 20% of the weighted Euclidean distances. The features and weights are taken from the 2014 Car-Brand Perception Survey of Consumer Reports [48].

TABLE 5. CAR FEATURES AND THEIR WEIGHTS

| Quality | Safety | Value | Perf | Eco | Design | Tech |
|---|---|---|---|---|---|---|
| 90 | 88 | 83 | 82 | 81 | 70 | 68 |

Out of the seven features in Table 5, we consider three scenarios of the known feature set: Scenario #1 - {Quality, Safety, Value, Perf}, Scenario #2 - {Quality, Safety, Value, Perf, Eco}, and Scenario #3 - {Quality, Safety, Value, Perf, Eco, Design}. We also assume that efforts have been made to estimate the consumers' perception of the values of the known features for the $N$ car brands. To simulate the estimates, we add noises to the generated values of the known features for the $N$ car brands. These noises follow normal distributions with zero means and standard deviations equal to 5% of the generated values of the corresponding known features.

As with the example in Section 4.3, conditional MDS is compared with metric MDS, ISOMAP, t-SNE, and UMAP in terms of the ACCs in this example. However, the ACCs here are calculated between the reduced-dimension space of all methods (including conditional MDS) and the true values of all seven features in Table 5 (i.e., before adding noises). Table 6 reports the medians (and the interquartile ranges in the parentheses) of the ACCs over 100 Monte Carlo replicates for all methods. Table 6 shows that conditional MDS improves the learning accuracy of the reduced-dimension space. The more known features are utilized, the better the learning accuracy is.

TABLE 6. MEDIAN (INTERQUARTILE RANGE) OF AVERAGE CANONICAL CORRELATIONS BETWEEN THE UNKNOWN FEATURES AND THE REDUCED-DIMENSION SPACE LEARNED FOR THE CAR BRANDS BY FOUR DIMENSION REDUCTION METHODS AND CONDITIONAL MDS OVER 100 REPLICATES.

| | | |
|---|---|---|
| Metric MDS | | .81 (.78, .84) |
| ISOMAP | | .74 (.71, .77) |
| t-SNE | | .72 (.69, .75) |
| UMAP | | .70 (.68, .73) |
| Conditional MDS | {Quality, Safety, Value, Perf} | .90 (.86, .93) |
| | {Quality, Safety, Value, Perf, Eco} | .94 (.92, .95) |
| | {Quality, Safety, Value, Perf, Eco, Design} | .97 (.96, .98) |

To provide more insights of conditional MDS, Table 7 reports the medians of the absolute correlations of the first conditional MDS dimension with each feature over the 100 Monte Carlo replicates (along with the interquartile ranges in the parentheses). Table 7 suggests similar observations with the previous examples. Particularly, the median absolute correlations are weak for the known features and generally strong for at least an unknown feature in all three scenarios. The weak median absolute correlations of the first conditional MDS dimension with the known features in Table 7 clearly indicate that conditional MDS effectively marginalizes the effects of the known features.

TABLE 7. MEDIAN (INTERQUARTILE RANGE) OF ABSOLUTE CORRELATIONS BETWEEN THE FIRST CONDITIONAL MDS DIMENSION AND EACH CAR FEATURE OVER 100 REPLICATES. THE BOLD NUMBERS INDICATE CORRELATIONS WITH UNKNOWN FEATURES.

| Scenario # | 1 | 2 | 3 |
|---|---|---|---|
| Quality | .11 (.05, .18) | .12 (.05, .17) | .12 (.05, .20) |
| Safety | .10 (.06, .15) | .10 (.04, .16) | .13 (.07, .24) |
| Value | .08 (.04, .17) | .12 (.06, .19) | .13 (.06, .20) |
| Perf | .10 (.06, .16) | .11 (.05, .18) | .11 (.05, .21) |
| Eco | **.91 (.79, .96)** | .10 (.04, .16) | .11 (.04, .21) |
| Design | **.27 (.16, .49)** | **.82 (.32, .93)** | .14 (.05, .24) |
| Tech | **.28 (.12, .51)** | **.74 (.41, .94)** | **.98 (.96, .98)** |

The results in Table 7 also demonstrate the advantage of conditional MDS in simplifying visualization and knowledge discovery tasks. Suppose a practitioner begins with the four known features in Scenario #1 and discover the Eco feature via visualizing the learned conditional MDS dimensions. The practitioner can then repeat this process with these four known features + Eco (i.e., Scenario

#2) and discover the Design feature. Repeating this process one more time with the four original known features + Eco + Design (i.e., Scenario #3), the practitioner can discover the Tech feature. This process is illustrated in Fig. S3 in the supplementary materials of this paper. Note that without conditional MDS, the practitioner would need to discover Eco, Design, and Tech together with rediscovering the other four known features at once; this is a much more challenging task.

Because the features in Table 5 contribute separately to the pairwise dissimilarities, the ground truth of **B** for each scenario in this example is a diagonal matrix. The diagonal values of this matrix are the weights of the corresponding known features. However, we assume this is unknown and have used the update (8) instead of (10) for **B** in this example. To demonstrate that conditional MDS can still learn well **B**, Fig. 2 shows the mean squared error (MSE) $\times 10^{-5}$ of the estimate of each element in **B** over 100 Monte Carlo replicates for the three aforementioned scenarios of known features. We can see from Fig. 2 that conditional MDS estimated **B** with high accuracy in all scenarios.

| Scenario #1 | | | |
|---|---|---|---|
| 71 | 16 | 16 | 20 |
| 18 | 57 | 16 | 16 |
| 16 | 16 | 46 | 11 |
| 18 | 16 | 10 | 55 |

| Scenario #2 | | | | |
|---|---|---|---|---|
| 47 | 12 | 17 | 11 | 15 |
| 16 | 32 | 15 | 14 | 15 |
| 14 | 16 | 29 | 10 | 11 |
| 11 | 12 | 12 | 30 | 17 |
| 17 | 13 | 9 | 14 | 30 |

| Scenario #3 | | | | | |
|---|---|---|---|---|---|
| 17 | 9 | 10 | 10 | 10 | 10 |
| 8 | 20 | 7 | 10 | 8 | 10 |
| 9 | 7 | 19 | 10 | 8 | 13 |
| 9 | 12 | 11 | 20 | 9 | 11 |
| 9 | 7 | 8 | 9 | 14 | 9 |
| 10 | 10 | 12 | 12 | 10 | 24 |

Fig. 2 Each cell shows the mean squared error ($\times 10^{-5}$) over 100 Monte Carlo replicates of the estimate for each element in **B** (at the corresponding location) by conditional SMACOF for the car-brand perception example, for three scenarios of the known feature set.

## 4.5. Cylinder Machining

This example involves surface measurements of 90 cylinders [49]. Specifically, the radial deviations from the norm are measured on a 210×64 grid for each cylinder surface. Hence, each cylinder corresponds to a 13,330-dimensional vector. Prior to the measurements, the cylinders are machined using nine different combinations of two lathe-turning parameters. The first parameter is turning speed, with values in {80, 70, 65} $m/min$. The second parameter is cutting depth, with values in {0.4, .08, 1.2} $mm$. There are 10 cylinders in each combination.

Suppose that the Euclidean distances between the cylinders are given as their pairwise dissimilarities. Conditional MDS is applied to these dissimilarities, using the two lathe-turning parameters as the known features. The goal is to identify potential unknown features that cause surface variation among the cylinders in this cylinder machining process. Using again the scree-plot-based approach in Section 5.2, $p = 3$ is selected as the number of unknown features (see Fig. S4 in the supplementary materials of this paper for the scree plot of this example).

Table 8 shows the absolute correlations between the lathe-turning parameters and the three conditional MDS dimensions. If the lathe-turning parameters are marginalized completely, we could expect that these correlations are close to 0. However, some of them are moderate. This suggests that the cylinders might not be machined exactly at the set values of the two lathe-turning parameters. Hence, conditional MDS can help the process controller realize that there are potentially unknown factors (e.g., voltage fluctuation or tool wear) causing variation in the two lathe-turning parameters of this machining process.

TABLE 8. ABSOLUTE CORRELATIONS OF THE TWO LATHE-TURNING PARAMETERS AND THREE CONDITIONAL MDS DIMENSIONS USING THESE PARAMETERS AS THE KNOWN FEATURES.

| | $u_{\bullet,1}$ | $u_{\bullet,2}$ | $u_{\bullet,3}$ |
|---|---|---|---|
| Turning speed | 0.03 | 0.41 | 0.02 |
| Cutting depth | 0.54 | 0.00 | 0.35 |

Recall that the optimal $p$ value is identified as 3. Therefore, there is another potential unknown feature, other than the two related to the variations in the two lathe-turning parameters. To investigate this, Fig. 3 plots $u_{\bullet,2}$ versus $u_{\bullet,1}$, with the top views of some cylinders superimposed at the corresponding values of $u_{\bullet,1}$ and $u_{\bullet,2}$. Note that these top views are magnified to show the small radial deviations from the cylinder norm. In general, the top cross-sections of the cylinders are wider than the bottom cross-sections (probably because the cylinders were fixed at the bottom while being machined).

It can be seen from Fig. 3 that the top cross-sections of the cylinders are ellipsoidal. In adidition, the angles of their orientations tend to increase in magnitude from the bottom-left to top-right corners. To see this change more clearly, the estimated orientation angles are shown in the middle of the top views. Each estimate is the median of the rotation angles of the 10 top cross-sections of each cylinder. The absolute values of these angle estimates agree with our above observation. Therefore, conditional MDS can help the process controller identify the third unknown feature, which corresponds to rotation of the cylinders around their axis.

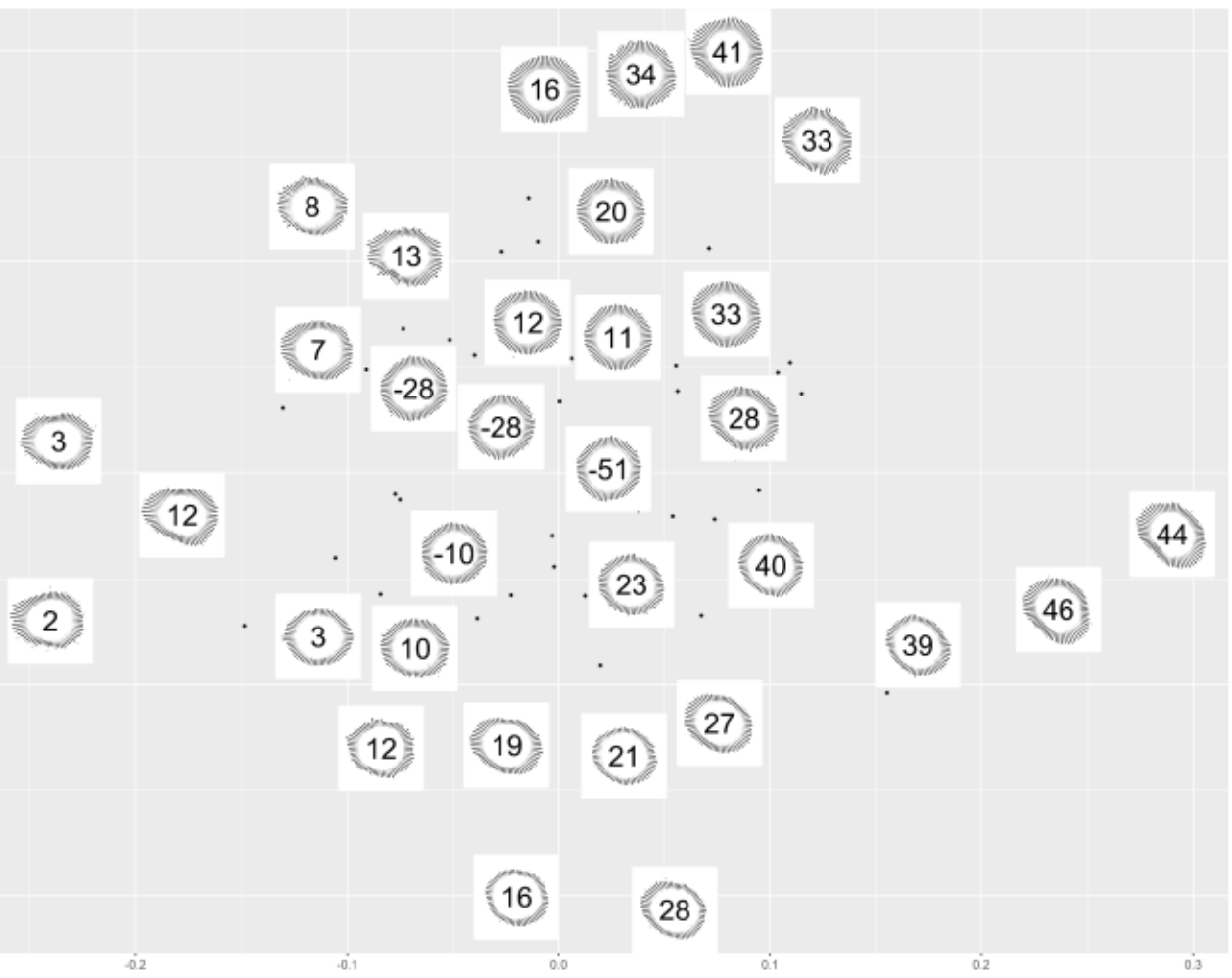

Fig. 3. Scatterplot of the two conditional MDS coordinates of the cylinders, using the two lathe-turning parameters as the known features. The top views of some cylinders are superimposed on the plot at the corresponding coordinates. The numbers in the top views are the estimated rotation angles of the cylinders around their axis.

# 5 Discussions

## 5.1. Validity of Assumptions

This section discusses the validity of the assumptions in Assumption 1, which is required for the convergence of conditional SMACOF. First, Assumption 1(a) virtually always holds because even if the dissimilarities $\delta_{ij}$ $(i, j = 1, 2, \dots, N)$ are not symmetric (which is uncommon in practice), we can make them symmetric by replacing $\delta_{ij}$ and $\delta_{ji}$ with $\delta_{ij} + \delta_{ji}$ $(i, j = 1, 2, \dots, N)$. And for symmetric dissimilarities, there are no reasons for using $w_{ij} \neq w_{ji}$ $(i, j = 1, 2, \dots, N)$. Second, Assumption 1(b) causes no loss of generality for conditional SMACOF. If this assumption is violated, we can apply conditional SMACOF to each group. By definition, the weight matrix of each group is irreducible [18], which satisfies Assumption 1(b). Third, Assumption 1(c) is violated if $\mathbf{V}$ contains less than $q$ linearly independent difference vectors $\mathbf{v}_j - \mathbf{v}_k$ $(1 \leq j < k \leq N)$. This situation is rather rare in practice because the number of the difference vectors grows quadratically with $N$. Reducing $q$ is an option to address this situation should it occurs. Alternatively, we can perturb $\mathbf{V}$ slightly such that it contains $q$ linearly independent difference vectors.

## 5.2. Choice of *p*

A natural question is how to choose the smallest number of unknown features $p$ without losing information in the data. The answer is similar to the case of conventional dimension reduction methods, for which the common solution is based on a scree plot of some measure of goodness-of-fit of the dimension reduction result. For example, the eigenvalues are often used in classical MDS, and the stress is commonly used in metric MDS for this purpose. For conditional MDS, we can choose $p$ at the elbow of the scree plot of the conditional stress versus $p$.

Fig. 4 shows such scree plots for the car-brand perception example in Section 4.4, using seven different sets of known features: {Quality} ($q$ = 1), {Quality, Safety} ($q$ = 2), {Quality, Safety, Value} ($q$ = 3), {Quality, Safety, Value, Perf} ($q$ = 4), {Quality, Safety, Value, Perf, Eco} ($q$ = 5), {Quality, Safety, Value, Perf, Eco, Design} ($q$ = 6), and {Quality, Safety, Value, Perf, Eco, Design, Tech} ($q$ = 7). Note that the solutions for the case of $p$ = 0 can be obtained by modifying conditional SMACOF to only update $\mathbf{B}$ and set $\mathbf{u}$-related values to 0. Fig. 4 shows that the elbow of each scree plot occurs at the ground truth of $p$, which is 7 – $q$. When $p$ is truly 0 (the case of $q$ = 7), the scree plot does not have a clear elbow. This is how we can determine if there exists any unknown feature in the data.

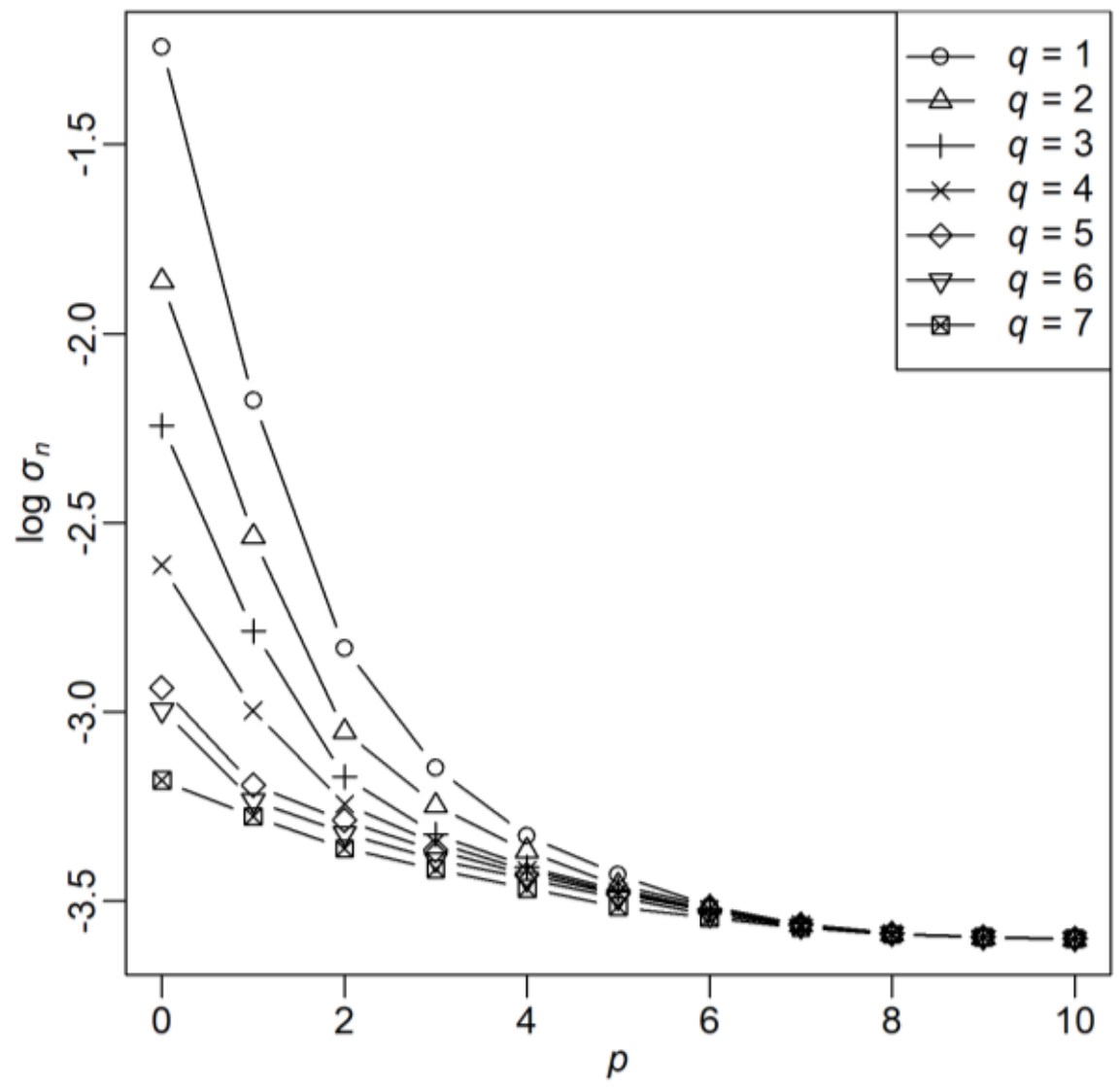

Fig. 4. Scree plots of normalized conditional stress $\sigma_n$ (in log scale) against different choices of the number of unknown features $p$, for seven sets of $q$ known features in the car-brand perception example.

## 5.3. Irrelevant Known Features

This section discusses the robustness of conditional MDS when irrelevant features are included in the known feature set. We use the car-brand perception simulation example in Section 4.4 for illustration. For brevity, we only consider Scenario #1 in this section. This means that Quality, Safety, Value, and Perf are the relevant known features. To create irrelevant known features, we generate a number of features in the same manner with those in Table 5, except that we do not use them in computing the dissimilarities between the car brands. We use $N$ = 100 in this example. We apply conditional MDS to this dataset, using Quality, Safety, Value, Perf, and the irrelevant features as the known features. The conditional SMACOF version for strictly diagonal $\mathbf{B}$ is used here, to focus entirely on the estimation quality of the diagonal elements $b_m$'s of $\mathbf{B}$.

Denote $\hat{b}_m$ as the estimate of $b_m$, and $s_m$ as the sample standard deviation of the $m$th column of $\mathbf{V}$ ($m$ = 1, 2, …, $q$). Let $\mathfrak{R}$ be the index set of the relevant features and $|\mathfrak{R}|$ be its cardinality. We evaluate the results of conditional MDS in this example based on the following two measures. The first measure is the MSE of the estimates of the $b_m$'s corresponding to the relevant features: $\sum_{m\in\mathfrak{R}}\left(\hat{b}_m - b_m\right)^2/|\mathfrak{R}|$. This measure tells us how accurately conditional MDS learns the effects of the relevant known features. The second measure is $r = \frac{\sum_{m\in\mathfrak{R}}|\hat{b}_m|s_m/|\mathfrak{R}|}{\sum_{m\in\{1,\dots,q\}\setminus\mathfrak{R}}|\hat{b}_m|s_m/(q-|\mathfrak{R}|)}$, which can be viewed as a "signal-to-noise" ratio. This measure is the average estimate of the $b_m$'s of the relevant features over that of the irrelevant features. The larger $r$ is, the better the accuracy of conditional MDS is. Note that $s_m$ appears in this measure to account for the differences in the feature scales.

Fig. 5 plots these two measures against the number of irrelevant features. We can see that the MSE measure slightly increases as the number of irrelevant features increases up to around 30, but seems to stabilize afterwards. More importantly, the MSEs are very small, even when the number of irrelevant features is close to the number of car brands $N$ = 100. This shows that conditional MDS manages to learn well the effects of the relevant known features. We can also see that the "signal-to-noise" ratio $r$ keeps increasing as the number of irrelevant features increases. Because the $\hat{b}_m$'s of the relevant features remain close to the ground truths (due to the small MSEs), the $\hat{b}_m$'s of the irrelevant features approach 0 as their number increases.

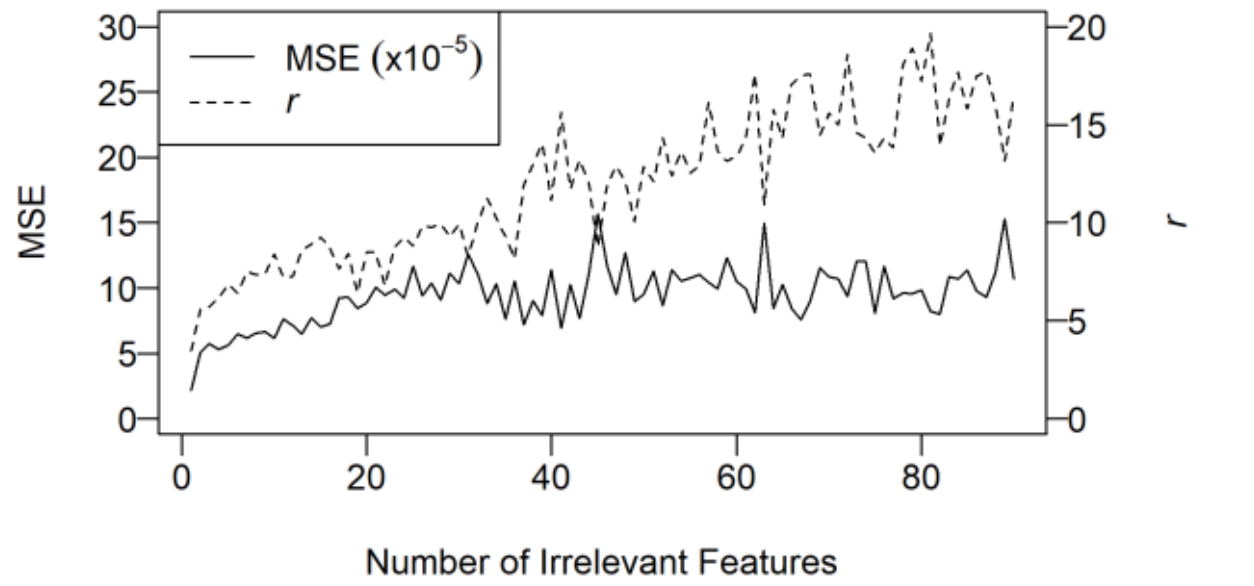

Fig. 5. Accuracy of conditional MDS in estimating **B** against the number of irrelevant features in the known feature set.

## 5.4. Interpretation with B

This section discusses two interpretations of **B** (see Section S2 in the supplementary materials of this paper for an illustrative example). The first involves the columns of **B**, where we assume that the known features are standardized. Recall that the known features are linearly combined with each column of **B**, creating affine-transformed features. Thus, we can use the values of each column of **B** to determine how the known features interrelate in contributing to the pairwise distances in (3). The second involves the rows of **B**, where we do not require standardized known features, but we assume that the dissimilarity $\delta_{ij}$ between the $i$th and the $j$th objects equals to $d_{ij}(\mathbf{U}, \mathbf{B})$ plus some noise. From (3), we see that an increase of one unit in the difference in the $m$th known feature between the $i$th and the $j$th objects will contribute an increase to the squared dissimilarity $\delta_{ij}^2$ by the sum of square of the elements in the $m$th row of **B**. Thus, the importance of a known feature to the dissimilarities between objects is reflected in the magnitudes of the elements in its corresponding row in **B**. In the special case when **B** is strictly diagonal, the sum of square of the $m$th row of **B** reduces to $b_m^2$. We encounter this situation in the example in Section 5.3, where the relevant known features have large $\hat{b}_m^2$'s, whereas the irrelevant features have $\hat{b}_m^2$'s close to 0.

## 5.5. Normalized Conditional Stress

This section illustrates the behavior of the normalized conditional stress in the examples used in this paper. Particularly, Fig. 6 plots the normalized conditional stress (in log scale) against the first 100 iterations of conditional MDS for each example. Obviously, the normalized conditional stress reduces significantly in the first few iterations (especially the first one) in all the examples. Then, it decreases (as expected from Theorem 1) gradually. In general, the normalized conditional stresses reach small values relatively quickly. These observations show that conditional SMACOF can be stopped quite early, at least in these examples. Indeed, this early stopping strategy is also widely used for the conventional SMACOF algorithm.

## 5.6. Computational Complexity

This section discusses the computational complexity of conditional SMACOF. The space complexity of conditional SMACOF is $O(N^2)$, and this applies to any algorithm that deals with dissimilarity matrices. Following, we discuss the time complexity of conditional SMACOF and compare it with SMACOF, ISOMAP, t-SNE, and UMAP.

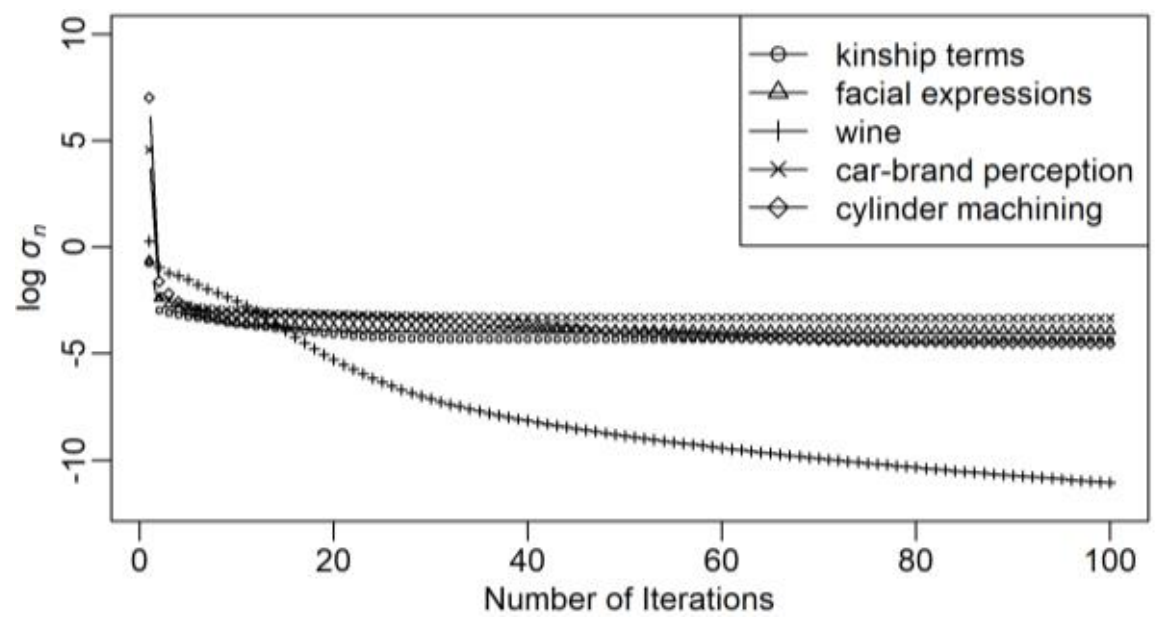

Fig. 6. Plots of normalized conditional stress $\sigma_n$ (in log scale) in the first 100 iterations of all examples.

Table S1 in the supplementary materials of this paper provides the time complexities of the main calculations in conditional SMACOF. If we assume that $N >> p + q$, the time complexity of conditional SMACOF can be derived from this table to be: (*i*) $O(N^2)$ when the $w_{ij}$'s are 1, and (*ii*) $O(N^{2.373})$ otherwise. Note that these time complexities are the same with those of the conventional SMACOF algorithm. The time complexities of ISOMAP and t-SNE are both $O(N^2)$ [19], [20], which are the same with that of conditional SMACOF when all the weights $w_{ij}$'s are 1. The time complexity of UMAP largely depends on its nearest neighbor search step [21], which usually have much better empirical performance than $O(N^2)$ if done approximately.

We now compare the computational times of conditional SMACOF (for both weighting schemes of the $w_{ij}$'s) with those of SMACOF, ISOMAP, t-SNE, and UMAP. Here we reuse the car-brand perception example in Section 4.4, but vary $N$ to study its impact on the computational time. We use {Quality, Safety, Value, Perf} as the known feature set for conditional MDS. Except for ISOMAP, which has an analytical solution, the other approaches rely on iterative optimization. Hence, we set the number of iterations for all these iterative methods at 100 to provide a clearer picture of their execution times. Fig. 7 plots the computational times of all methods. Clearly, t-SNE is the slowest method, especially for larger $N$. The execution times of SMACOF, ISOMAP, and conditional SMACOF are somewhat similar for the tested $N$ values. However, the execution time of ISOMAP appears to increase rapidly as $N$ increases.

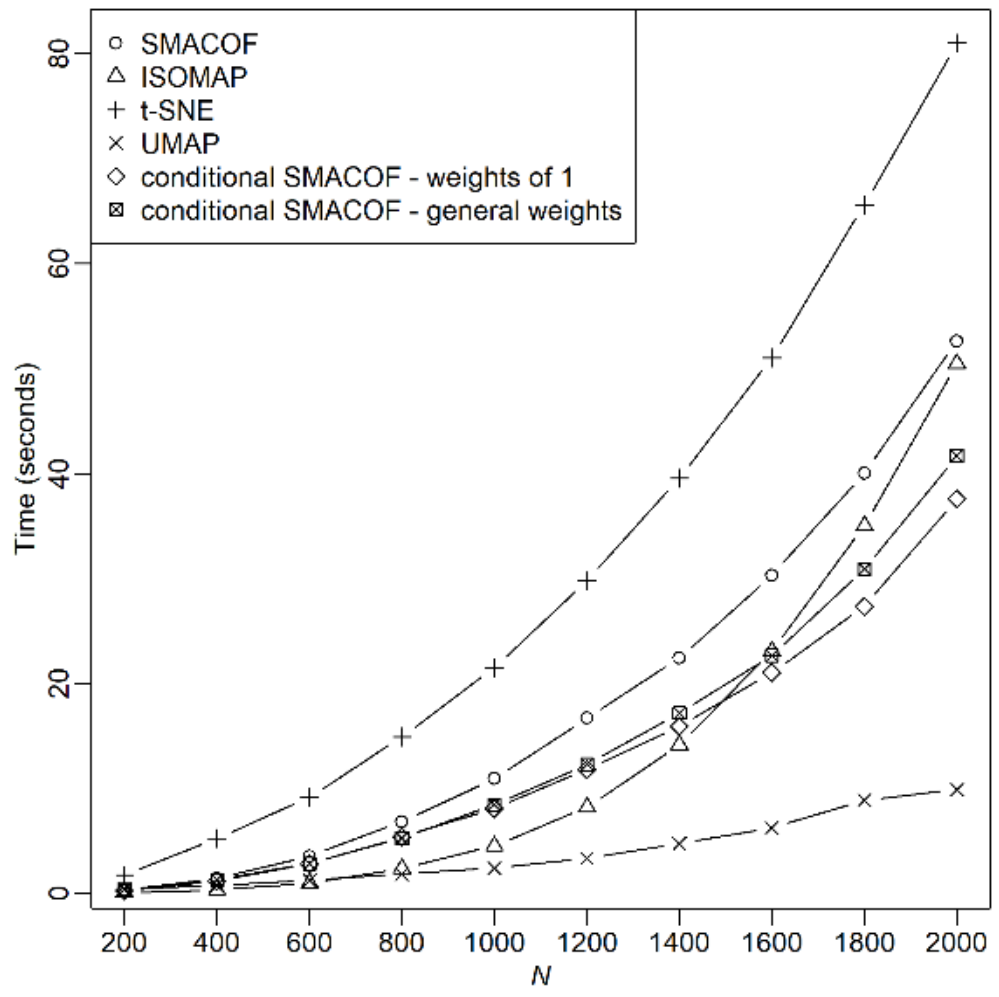

Fig. 7. Computational time comparison for SMACOF, ISOMAP, t-SNE, UMAP, and the proposed conditional SMACOF algorithm.

UMAP seems to be the fastest approach. Nevertheless, its ACC performance is much worse than that of conditional MDS and downgrades significantly as $N$ increases (see Fig. 8). Thus, UMAP would require much more time to execute enough iterations to obtain good dimension reduction results for larger $N$ values. For the example in Fig. 7 with $N$ = 2000, UMAP needs around 100 and 200 seconds to perform 5000 and 10000 iterations, which result in ACCs of 0.56 and 0.60, respectively. Nonetheless, these ACCs are still much poorer than that of conditional SMACOF in this example. The latter remains close to the max ACC for all the tested $N$ values, using just 100 iterations.

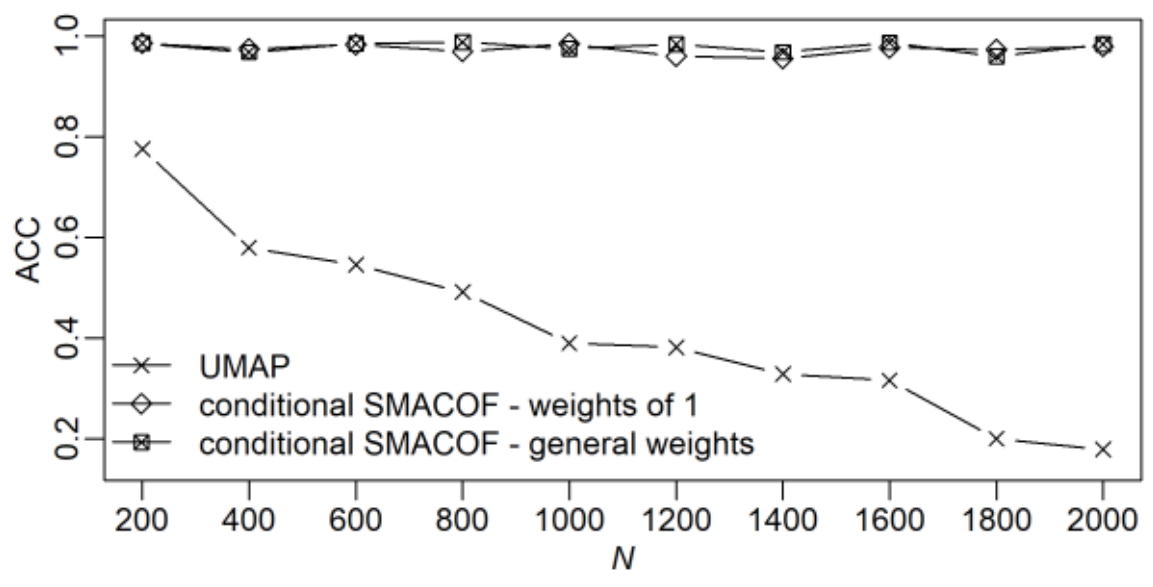

Fig. 8. Average canonical correlations (ACCs) of UMAP and conditional SMACOF corresponding to the example in Fig. 7.

## 6 Conclusions

The problem of mapping high-dimensional data to a low-dimensional space in the presence of other known features is ubiquitous in science and engineering. This is because there are often controllable/measurable features in most applications. To solve this problem, this paper proposes a broad class of conditional MDS methods. A conditional SMACOF algorithm is developed to optimize the conditional stress objective function of conditional MDS. This algorithm is shown to converge under mild assumptions.

Conditional MDS is illustrated with several examples, which demonstrate its advantages over conventional dimension reduction methods, due to its capability to utilize available knowledge of known features. This capability helps improve the estimation quality of the reduced-dimension space. Furthermore, this capability helps conditional MDS enhance knowledge discovery processes in two ways. First, conditional MDS exposes only unknown, unanticipated features in the reduced-dimension space; this helps avoid the aforementioned masking problems. Second, conditional MDS enables a repeated, simpler knowledge discovery process, in which discovered features in previous analyses can be used as the known features in subsequent analyses.

Conditional MDS is not without limitations. First, conditional MDS does not directly solve the challenges in interpreting the unknown features. Rather, it helps practitioners approach this task as accurately and easily as possible, by utilizing more fully existing knowledge. Methods that improve the interpretation of dimension reduction results [50] can be explored to address this limitation. Second, the time complexity of conditional MDS does not scale well with $N$. As there is a connection between SMACOF and gradient based optimization [18], stochastic gradient based optimization with an $O(N)$ time complexity may be a potential direction to address the scalability problem. These issues are left for future research.

Another potential research avenue is extending conditional MDS to other application contexts. For instance, a challenge in precipitation forecasting is that data come from nonhomogeneous and overlapped sources, such as satellites, radar, and other weather measurements. If the non-vector data can be transformed into pairwise dissimilarities, the learned conditional MDS features (using the vector data as the known features) can be used as inputs for conventional predictive models. Another example is handling missing data in, for instance, high-dimensional and sparse matrices [51] and dynamically weighted directed network [52]. Data in such cases can be treated as pairwise dissimilarities. Assuming that information of known features is available, conditional MDS can incorporate this knowledge to estimate coordinate representations for the objects with missing dissimilarities. The coordinates can in turn be used to estimate these missing values.

## Acknowledgements

The author thanks the editor and the anonymous associate editor and referees for helping improve this article.

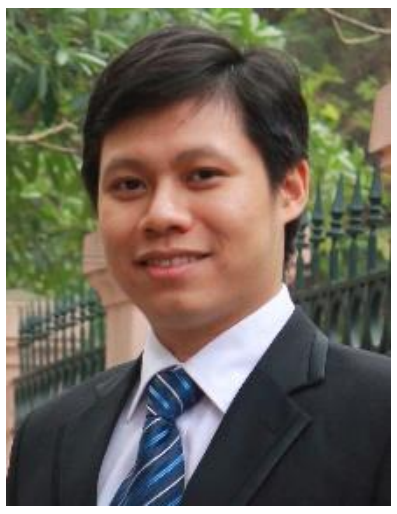

**Anh T. Bui** received the B.S. degree in electrical engineering from Hanoi University of Science and Technology, Vietnam, the M.S. degree in industrial and management engineering from Pohang University of Science and Technology, South Korea, and the Ph.D. degree in industrial engineering & management sciences from Northwestern University, USA. He is currently an assistant professor in the Department of Statistical Sciences & Operations Research at Virginia Commonwealth University, USA. His research interests lie in applied statistics and machine learning, for analyzing data from manufacturing, healthcare, and other enterprise systems. Dr. Bui is a recipient of the Lloyd S. Nelson Award from the American Society for Quality.

**Supplementary Materials for**

**“Dimension Reduction with Prior Information for Knowledge Discovery”**

Anh Tuan Bui

## S1. Figures and Tables

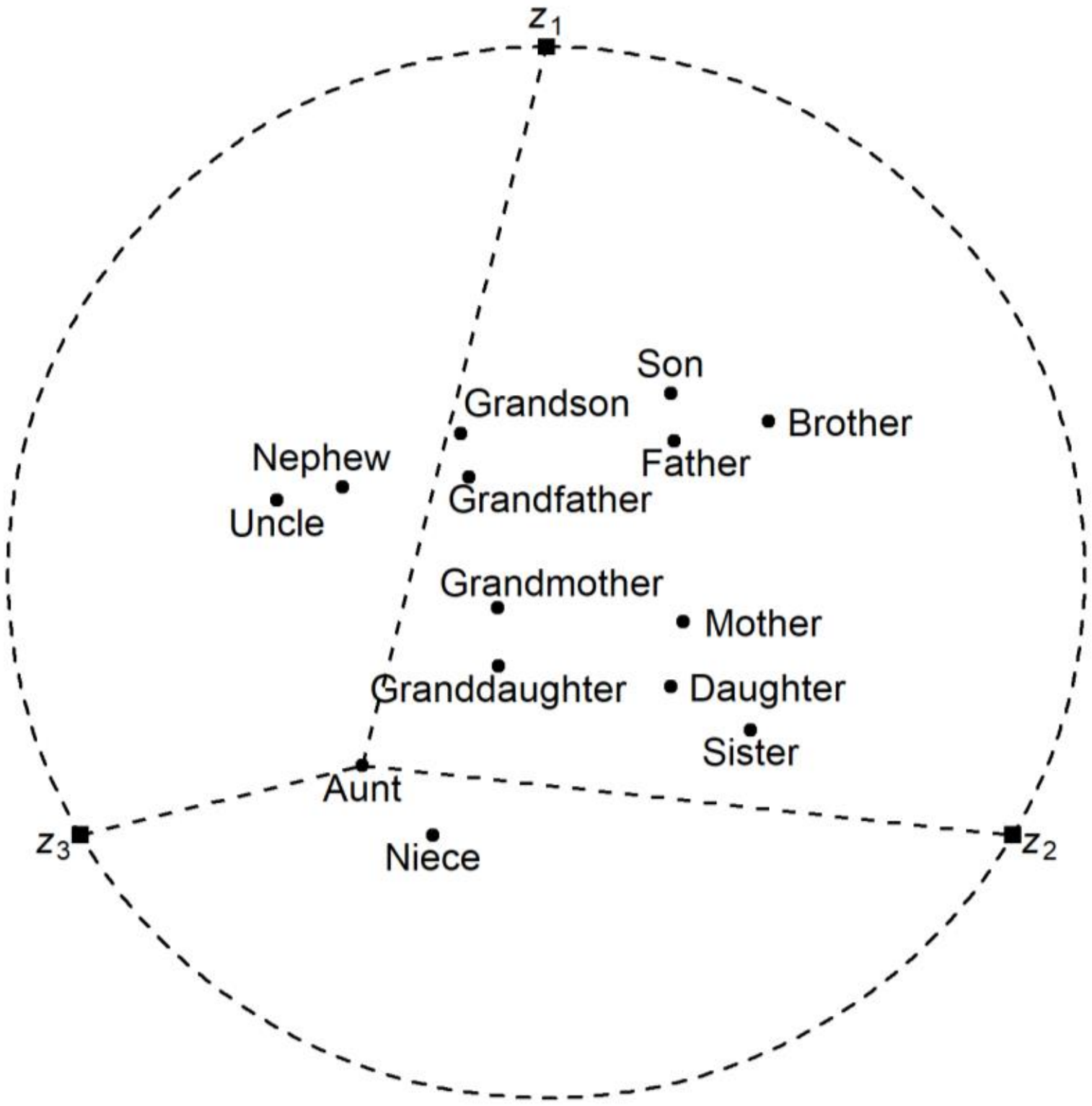

Fig. S1. RadViz plot of the three coordinates found by metric MDS for the 14 kinship terms in Fig. 1. This plot represents these three coordinates by anchors $z_1$, $z_2$, and $z_3$, which are equally spaced around a unit circle. Each point inside the circle (i.e., a kinship term) is positioned at the equilibrium of three imaginary springs that connect the point to the three anchors. The stiffness of each spring is proportional to the corresponding coordinate value, which has been rescaled to be in [0, 1]. Points that are closer to an anchor have higher values of the corresponding coordinate. For instance, the three dashed line represent the three springs attaching to the “Aunt” point, which has a large “$z_3$” coordinate value and small “$z_1$” and “$z_2$” coordinate values. This plot reveals the Gender feature (see the kinship terms in the upper versus lower halves) and the Kinship Degree features (see the horizontal transition in the kinship degree level of the kinship terms) together. However, the Generation feature could not be inferred from this plot.

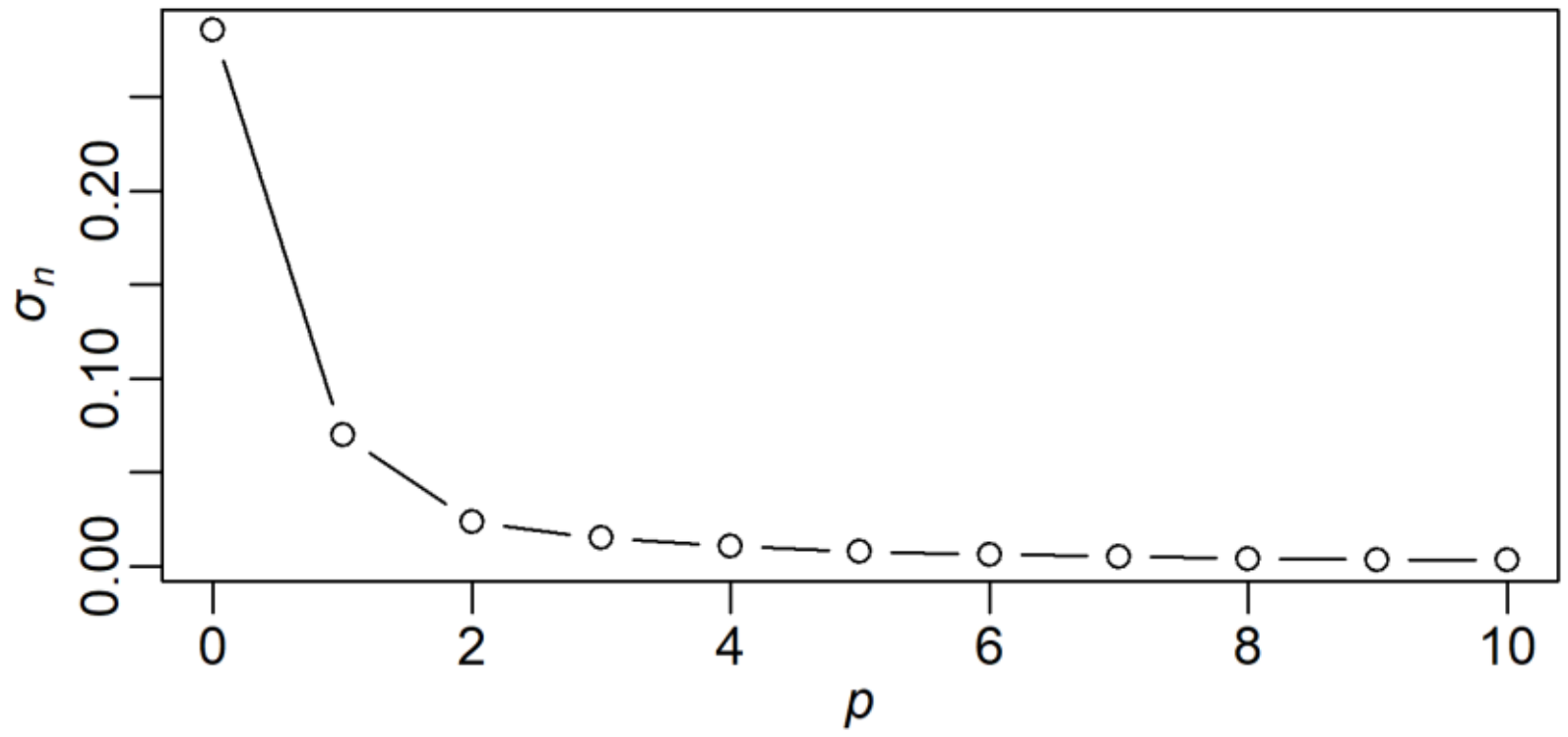

Fig. S2. Scree plot of normalized conditional stress $\sigma_n$ against different choices of the number of unknown features $p$ in the textile fabrics example.

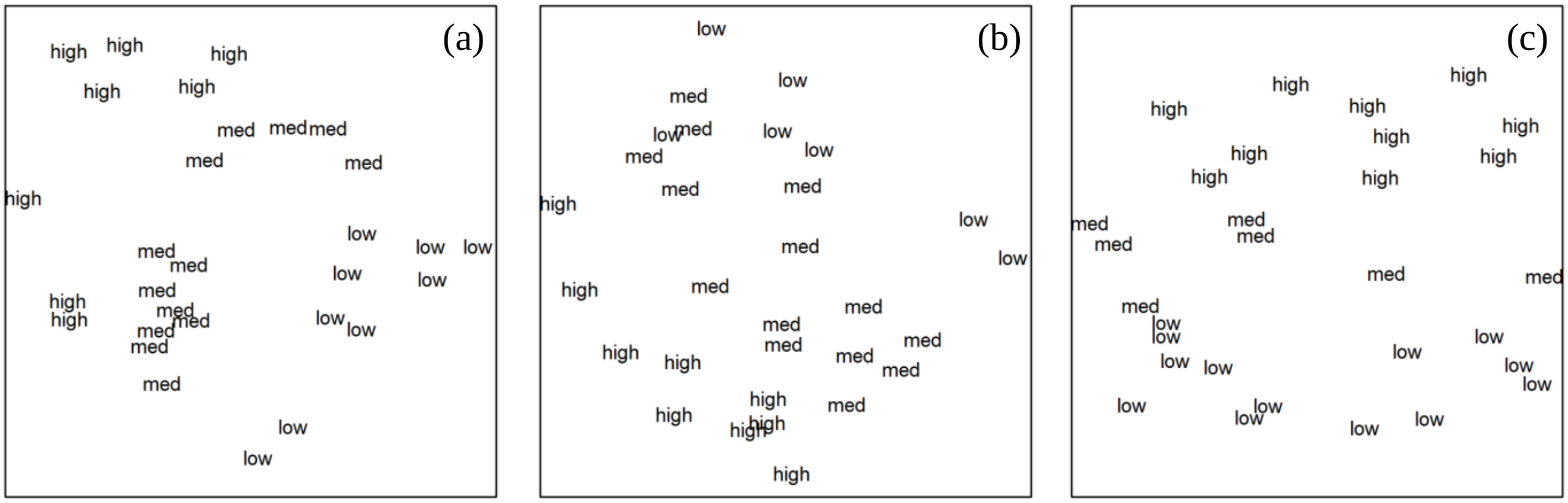

Fig. S3. Scatterplot of the two conditional MDS coordinates for the car brands, using the known feature set: (a) {Quality, Safety, Value, Perf}, (b) {Quality, Safety, Value, Perf, Eco}, and (c) {Quality, Safety, Value, Perf, Eco, Design}. The symbols "low", "med", and "high" indicate the car brand positions and correspond to low, medium, or high values, respectively, of: (a) Eco, (b) Design, and (c) Technology, which we assume the practitioners can relate to from their knowledge of the car brands.

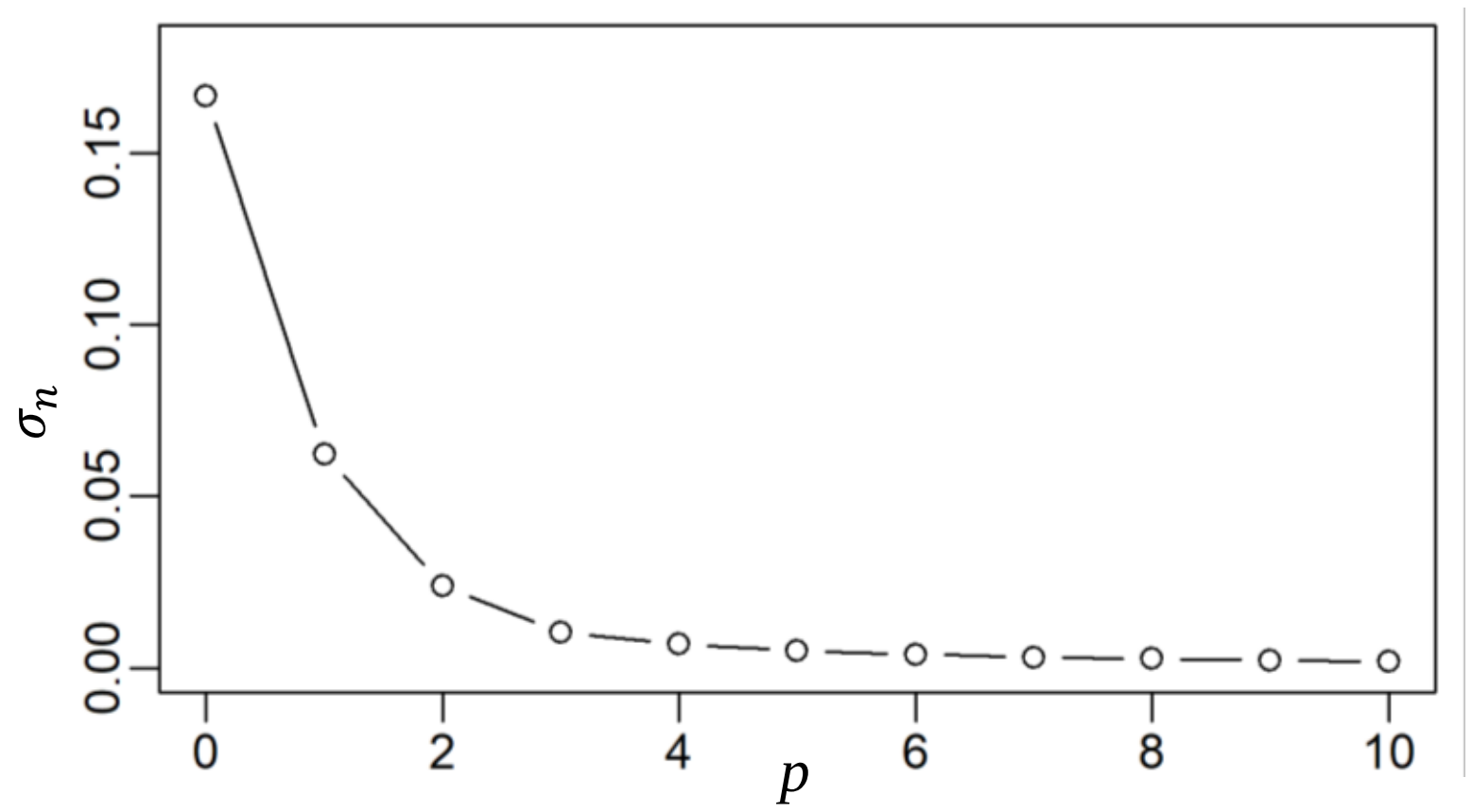

Fig. S4. Scree plot of normalized conditional stress $\sigma_n$ against different choices of the number of unknown features $p$ in the cylinder machining example.

Table S1. Time complexities of the main calculations in conditional SMACOF

| | Calculation | General weights | Weights of 1 | |
|---|---|---|---|---|
| | | | Exact | Large-$N$ approximation |
| Step 1 | $\mathbf{H}^+$ | $O(N^{2.373})$ | $O(N^2)$ | $O(1)$ |
| | $(\mathbf{V}^T\mathbf{H}\mathbf{V})^{-1}\mathbf{V}^T$ | $O(N^2q + q^2[2N + q^{0.373}])$ | | |
| Each iteration in Step 2 | $\mathbf{C}(\mathbf{U}, \mathbf{B})$ | $O(N^2[p + q])$ | | |
| | $\mathbf{U}$ | $O(N^2p)$ | | |
| | $\mathbf{B}$ | $O(N^2q + [N + q^{0.373}]q^2)$ | | |
| | $\sigma_n(\mathbf{U}, \mathbf{B})$ | $O(N^2)$ | | |

Table S2. Estimates of **B** in the facial expression example, using PU, AR, and TS as the known features.

| | Standardized Known Features? | | | | | |
|---|---|---|---|---|---|---|
| | Yes | | | No | | |
| PU | 1.03 | -0.03 | 0.02 | 0.43 | -0.01 | 0.01 |
| AR | -0.06 | 0.85 | -0.26 | -0.02 | 0.35 | -0.12 |
| TS | 0.03 | -0.20 | 0.95 | 0.00 | -0.07 | 0.43 |

## S2. Interpretation with B

This section illustrates how we can use the estimate of **B** for interpretation. For the first interpretation, the 3×3 cells corresponding to the standardized case in Table S2 show the estimate of **B** for the facial expression example (in Section 4.1), using all three standardized Schlosberg scales as the known features. As mentioned above, the three Schlosberg scales are linearly combined to create three affine-transformed features. The first feature is 1.03×PU – 0.06×AR + 0.03×TS, which mainly involves PU; AR and TS have negligible impact on this feature. The second feature is -0.03×PU + 0.85×AR – 0.20×TS, which mainly involves AR and, to a lesser extent and in opposite direction, TS. The third feature is 0.02×PU – 0.26×AR + 0.95×TS, which is similar to the second feature, but with AR and TS swapping the roles. For both second and third features, PU has almost no influence.

For the second interpretation, the 3×3 cells corresponding to the non-standardized case in Table S2 show the estimate of **B** for the facial expression example (in Section 4.1), using all three original Schlosberg scales as the known features. In this example, PU and TS contribute the most to the difference between facial expressions. AR also contributes to the difference between facial expressions, but to a lesser extent.

## S3. Proofs

### S3.1. Lemma 1

We have:

$$\mathbf{a}^T\mathbf{V}^T\mathbf{H}\mathbf{V}\mathbf{a} = \mathbf{a}^T\left[\sum_{j<k} w_{jk}(\mathbf{v}_j - \mathbf{v}_k)(\mathbf{v}_j - \mathbf{v}_k)^T\right]\mathbf{a}$$

$$= \sum_{j<k} w_{jk}\mathbf{a}^T(\mathbf{v}_j - \mathbf{v}_k)(\mathbf{v}_j - \mathbf{v}_k)^T\mathbf{a}$$

$$= \sum_{j<k} w_{jk}[\mathbf{a}^T(\mathbf{v}_j - \mathbf{v}_k)]^2.$$

Because the weights are non-negative, $\mathbf{a}^T\mathbf{V}^T\mathbf{H}\mathbf{V}\mathbf{a} \geq 0$. Thus, $\mathbf{V}^T\mathbf{H}\mathbf{V}$ is positive semi-definite. Under Assumption 1(b), there exists at least a positive $w_{jk}$ for each object $j$ = 1, 2, …, $N$ and an object $k > j$. Assumption 1(c) implies that we can select some $q$ pairs of ($j$, $k$)'s such that $\sum_{m=1}^{q} a_m(v_{jm} - v_{km}) = 0$ if and only if $[a_1, a_2, \dots, a_q]^T = \mathbf{0}$. Hence, $\mathbf{V}^T\mathbf{H}\mathbf{V}$ is positive definite■

### S3.2. Theorem 1

Assume that Assumption 1 holds and expand (2) to:

$$\sigma(\mathbf{U}, \mathbf{B}) = \eta_{\tilde{\delta}}^2 + \eta^2(\mathbf{U}, \mathbf{B}) - 2\rho(\mathbf{U}, \mathbf{B}), \qquad \text{(S1)}$$

where $\eta_{\tilde{\delta}}^2 = \sum_{i<j} w_{ij}\tilde{\delta}_{ij}^2$ is a constant, $\eta^2(\mathbf{U}, \mathbf{B}) = \sum_{i<j} w_{ij}d_{ij}^2(\mathbf{U}, \mathbf{B})$, and $\rho(\mathbf{U}, \mathbf{B}) =$ $\sum_{i<j} w_{ij}\tilde{\delta}_{ij}d_{ij}(\mathbf{U}, \mathbf{B})$. It is not difficult to verify that $\sum_{i<j} w_{ij}\|\mathbf{u}_i - \mathbf{u}_j\|^2 = \text{tr}(\mathbf{U}^T\mathbf{H}\mathbf{U})$. Similarly, we can show that $\sum_{i<j} w_{ij}\|\mathbf{B}^T(\mathbf{v}_i - \mathbf{v}_j)\|^2 = \text{tr}(\mathbf{B}^T\mathbf{V}^T\mathbf{H}\mathbf{V}\mathbf{B})$. Thus, the second summand in (S1) is:

$$\eta^2(\mathbf{U}, \mathbf{B}) = \sum_{i<j} w_{ij}\|\mathbf{u}_i - \mathbf{u}_j\|^2 + \sum_{i<j} w_{ij}\|\mathbf{B}^T(\mathbf{v}_i - \mathbf{v}_j)\|^2$$

$$= \text{tr}(\mathbf{U}^T\mathbf{H}\mathbf{U}) + \text{tr}(\mathbf{B}^T\mathbf{V}^T\mathbf{H}\mathbf{V}\mathbf{B}). \qquad \text{(S2)}$$

The third summand in (S1) can also be shown to be $-2\rho(\mathbf{U}, \mathbf{B}) = -2\text{tr}([\mathbf{U},\ \mathbf{V}\mathbf{B}]^T\mathbf{C}(\mathbf{U}, \mathbf{B})[\mathbf{U},$

$\mathbf{VB}]$). Let $\mathbf{Z}_u \in \mathbb{R}^{N\times p}$ and $\mathbf{Z}_b \in \mathbb{R}^{q\times q}$ be two matrices having the same shapes with $\mathbf{U}$ and $\mathbf{B}$, respectively. Using the Cauchy-Schwarz inequality, it can be shown that:

$$-2\rho(\mathbf{U},\mathbf{B}) \leq -2\mathrm{tr}([\mathbf{U},\ \mathbf{VB}]^T\mathbf{C}(\mathbf{Z}_u,\mathbf{Z}_b)[\mathbf{Z}_u,\ \mathbf{VZ}_b])$$

$$= -2\mathrm{tr}(\mathbf{U}^T\mathbf{C}(\mathbf{Z}_u,\mathbf{Z}_b)\mathbf{Z}_u) - 2\mathrm{tr}(\mathbf{B}^T\mathbf{V}^T\mathbf{C}(\mathbf{Z}_u,\mathbf{Z}_b)\mathbf{VZ}_b), \tag{S3}$$

and the equality occurs when $\mathbf{Z}_u = \mathbf{U}$ and $\mathbf{Z}_b = \mathbf{B}$. Plugging (S2) and (S3) into (S1), we have:

$$\sigma(\mathbf{U},\mathbf{B}) \leq \eta_{\tilde{\delta}}^2 + \mathrm{tr}\mathbf{U}^T\mathbf{HU} + \mathrm{tr}\mathbf{B}^T\mathbf{V}^T\mathbf{HVB} - 2\mathrm{tr}(\mathbf{U}^T\mathbf{C}(\mathbf{Z}_u,\mathbf{Z}_b)\mathbf{Z}_u) - 2\mathrm{tr}(\mathbf{B}^T\mathbf{V}^T\mathbf{C}(\mathbf{Z}_u,\mathbf{Z}_b)\mathbf{VZ}_b)$$

$$= \tau(\mathbf{U},\mathbf{B},\mathbf{Z}_u,\mathbf{Z}_b),$$

which is a majorizing function of the conditional stress $\sigma(\mathbf{U},\mathbf{B})$. This function has a quadratic form in $\mathbf{U}$ and $\mathbf{B}$, and therefore, its global minimizer over $\mathbf{U}$ and $\mathbf{B}$ satisfies:

$$\begin{cases} \mathbf{0} = \frac{\partial\tau(\mathbf{U},\mathbf{B},\mathbf{Z}_u,\mathbf{Z}_b)}{\partial\mathbf{U}} \\ \mathbf{0} = \frac{\partial\tau(\mathbf{U},\mathbf{B},\mathbf{Z}_u,\mathbf{Z}_b)}{\partial\mathbf{B}} \end{cases} \leftrightarrow \begin{cases} \mathbf{HU} = \mathbf{C}(\mathbf{Z}_u,\mathbf{Z}_b)\mathbf{Z}_u \\ \mathbf{V}^T\mathbf{HVB} = \mathbf{V}^T\mathbf{C}(\mathbf{Z}_u,\mathbf{Z}_b)\mathbf{VZ}_b \end{cases}.$$

As a result, a minimizer of $\tau(\mathbf{U},\mathbf{B},\mathbf{Z}_u,\mathbf{Z}_b)$ over $\mathbf{U}$ can be obtained via the Guttman transform ([S1]):

$$\mathbf{U} = \mathbf{H}^+\mathbf{C}(\mathbf{Z}_u,\mathbf{Z}_b)\mathbf{Z}_u. \tag{S4}$$

And the minimizer of $\tau(\cdot)$ over $\mathbf{B}$ can be computed by

$$\mathbf{B} = (\mathbf{V}^T\mathbf{HV})^{-1}\mathbf{V}^T\mathbf{C}(\mathbf{Z}_u,\mathbf{Z}_b)\mathbf{VZ}_b, \tag{S5}$$

as $\mathbf{V}^T\mathbf{HV}$ is invertible (by Lemma 1).

Obviously, Eqs. 7 and 8 in Theorem 1 arise by setting $\mathbf{U} = \mathbf{U}^{[l]}$ and $\mathbf{Z}_u = \mathbf{U}^{[l-1]}$ in (S4) and $\mathbf{B} = \mathbf{B}^{[l-1]}$ and $\mathbf{Z}_b = \mathbf{B}^{[l-1]}$ in (S5). Based on the properties of majorization and minimization, we have the following inequality chain: $\sigma(\mathbf{U}^{[l]},\mathbf{B}^{[l]}) \leq \tau(\mathbf{U}^{[l]},\mathbf{B}^{[l]},\mathbf{U}^{[l-1]},\mathbf{B}^{[l-1]}) \leq \tau(\mathbf{U}^{[l-1]},\mathbf{B}^{[l-1]},\mathbf{U}^{[l-1]},\mathbf{B}^{[l-1]}) = \sigma(\mathbf{U}^{[l-1]},\mathbf{B}^{[l-1]})$, and the equality occurs if $\mathbf{U}^{[l]} = \mathbf{U}^{[l-1]}$ and $\mathbf{B}^{[l]} = \mathbf{B}^{[l-1]}$ ■

**S3.3. Corollary 1**

Theorem 1 implies that, under Assumption 1, updating $\mathbf{U}$ and $\mathbf{B}$ by (7) and (8), respectively, as in Algorithm 1 (with $l_{\max} = \infty$ and $\gamma = 0$) yields a strictly decreasing sequence of conditional stress values, until the conditional stress does not change. In other words, $\mathbf{U}^{[l]}$ and $\mathbf{B}^{[l]}$ will converge to $\mathbf{U}^*$ and $\mathbf{B}^*$ (respectively) as $l$ increases. We will show that $\mathbf{U}^*$ and $\mathbf{B}^*$ correspond to a critical point of the conditional stress function by contradiction.

Suppose on the contrary that $[[\mathbf{U}^*]^T, [\mathbf{B}^*]^T]^T$ is not a critical point of the conditional stress function. In this case, the gradient of the conditional stress function at $[[\mathbf{U}^*]^T, [\mathbf{B}^*]^T]^T$ exists and is non-zero. Then, there exists an arbitrarily small matrix $\mathbf{E}$ such that $\sigma(\mathbf{U}^* + \mathbf{E}, \mathbf{B}^* + \mathbf{E}) > \sigma(\mathbf{U}^*, \mathbf{B}^*)$. By the majorization property, $\tau(\mathbf{U}^* + \mathbf{E}, \mathbf{B}^* + \mathbf{E}) \geq \sigma(\mathbf{U}^* + \mathbf{E}, \mathbf{B}^* + \mathbf{E}) > \sigma(\mathbf{U}^*, \mathbf{B}^*) = \tau(\mathbf{U}^*, \mathbf{B}^*)$, i.e., there exists a matrix $\mathbf{E}$ with an arbitrarily small magnitude such that $\tau(\mathbf{U}^* + \mathbf{E}, \mathbf{B}^* + \mathbf{E}) > \tau(\mathbf{U}^*, \mathbf{B}^*)$. This contradicts the fact that the gradient of the majorizing function $\tau(\cdot)$ at $[[\mathbf{U}^*]^T, [\mathbf{B}^*]^T]^T$ is $\mathbf{0}$. Thus, $[[\mathbf{U}^*]^T, [\mathbf{B}^*]^T]^T$ is a critical point of the conditional stress function by contradiction■